\documentclass[11pt,a4paper,logo,nonumbering]{noematrix}
\usepackage[authoryear,sort&compress,round]{natbib}
\usepackage{ulem} % As in the template: \emph underlines venues in references.
\usepackage{amsmath,booktabs,multirow,graphicx,url,tikz,listings,colortbl,wrapfig}
\usepackage[section]{placeins}
\usepackage{newunicodechar}
\usepackage{capt-of}
\newunicodechar{，}{,}
\usetikzlibrary{arrows.meta,positioning,calc}

\usepackage{amsmath,amsfonts,bm}

\def\eqref#1{equation~\ref{#1}}

\def\1{\bm{1}}

\DeclareMathAlphabet{\mathsfit}{\encodingdefault}{\sfdefault}{m}{sl}
\SetMathAlphabet{\mathsfit}{bold}{\encodingdefault}{\sfdefault}{bx}{n}

\newif\ifshowtoc
\showtoctrue % Separate contents page; use \showtocfalse to omit it.
\definecolor{rsirow}{HTML}{EAF4F0}
\newcommand{\gain}[1]{{\color[HTML]{7D848C}\scriptsize\,(#1)}}

\hypersetup{pdftitle={RoboRSI: Stable, Efficient, and Reusable Robot Self-Evolution in Complex Real-World Environments},pdfauthor={Zimo Wen, Yijin Chen, Yuxuan Cao, Wendi Chen, Yanwen Zou, Wenye Yu, Fuhang Kuang, Han Xue, Jun Lv, Chuan Wen, Cewu Lu}}

\title{\centering RoboRSI: Stable, Efficient, and Reusable Robot Self-Evolution in Complex Real-World Environments}
\author{Zimo~Wen$^{*}$,\hspace{0.9em}Yijin~Chen$^{*}$,\hspace{0.9em}Yuxuan~Cao$^{*}$,\hspace{0.9em}Wendi~Chen,\hspace{0.9em}Yanwen~Zou,\hspace{0.9em}Wenye~Yu,\hspace{0.9em}Fuhang~Kuang,\hspace{0.9em}Han~Xue,\hspace{0.9em}Jun~Lv,\hspace{0.9em}Chuan~Wen$^{\dagger}$,\hspace{0.9em}Cewu~Lu$^{\dagger}$}
\correspondingauthor={Equal contribution.\quad $^{\dagger}$Corresponding authors.}
\begin{abstract}
A generalist robot in complex real-world environments should not only
perform diverse tasks but also improve through experience, turning what it
learns during execution into capabilities that later tasks can reuse. Robot agents that act through code
can already repair programs from execution feedback, yet it remains a
central challenge to organize this experience around the task structure
that gives it meaning, so that each repair is attributed to the responsible
capability, supported by execution evidence, and validated before it is
reused. We introduce RoboRSI, a robot self-improvement system built on
Top-Down Skill Refinement (TSR). TSR decomposes tasks into compound, atomic,
and base skills with scoped responsibilities and explicit input--output
contracts, attributes each execution outcome to the responsible branch, and
confines revision to that branch. Building upon this structure, a Manager,
Planner, Engineer, and Reviewer coordinate planning, execution, diagnosis,
and the validated release of new skills, while people steer the process
through objectives and corrections; stable skill sequences are further
consolidated into reusable compound skills. On a mobile manipulator, RoboRSI
develops multi-object household cleanup over 104 rounds. In simulation, it
achieves the highest success rate on LIBERO, LIBERO-PRO, LIBERO-Plus, and
RoboTwin, exceeding the strongest baseline by 2.7 to 11.0 percentage points.

\vspace{4pt}\noindent\textbf{Project page:} \url{https://lab.noematrix.ai/blog/2-roborsi-research-preview/}\quad
\textbf{Code:} \url{https://github.com/nssmd/RoboRSI}
\end{abstract}

\begin{document}
\maketitle

\noindent\begin{minipage}{\linewidth}
\centering
\vspace{2pt}
\includegraphics[width=.57\linewidth]{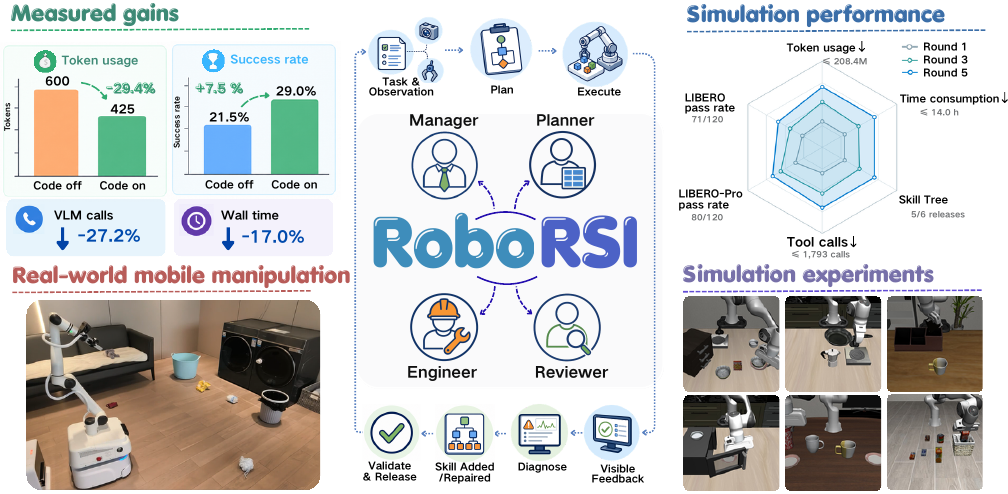}
\captionof{figure}{\textbf{RoboRSI at a glance.} The multi-agent self-improvement loop (center), real-world mobile manipulation and compound-skill gains (left), and simulation results (right).}
\label{fig:teaser}
\end{minipage}

\ifshowtoc
\clearpage
\tableofcontents
\clearpage
\fi

\section{Introduction}
\label{sec:intro}
The vision of generalist robotics is to develop robots that perform a broad range of tasks reliably in complex real-world environments, where objects, layouts, and instructions keep changing, while improving through experience. This shifts the goal of robot agents from episode-level task completion toward sustained capability accumulation: failed grasps, stale object locations, and recovery situations are valuable only if the robot can identify the capability involved, refine it using execution evidence, and make the resulting behavior available to later tasks.

A promising route lets foundation models act through code. Code as Policies~\citep{liang2023code} composes perception and control primitives into executable robot programs, and CaP-X~\citep{fu2026capx} studies how abstraction, grounding, and interaction shape such programs. Subsequent work closes the loop with execution feedback: ETA/OpenETA~\citep{chen2026eta} interleaves planning and tool execution with fresh observations, ASPIRE~\citep{lu2026aspire} adds repaired programs to a growing skill library, and ENPIRE~\citep{xiao2026enpire} carries this loop into physical experimentation. Execution feedback thus shifts from a signal for correcting a single attempt to a resource for building capabilities that persist across tasks.

Execution experience, however, becomes a reusable capability only when it is organized around the task structure that gives it meaning. A repair must remain connected to the objective that motivated it, belong to the capability responsible for the observed behavior, and be validated before it is reused. Without such structure, repeated attempts fill the context with complete trajectories, local fixes drift from the overall objective, and people are drawn back into resetting environments, reading logs, and deciding where each fix belongs.

Our observation is that this structure must serve both the people directing the robot and the agents implementing its behavior, which calls for two complementary abstractions. \emph{Human-friendly steering} lets people express objectives, domain knowledge, constraints, and safety judgments at the level of the service they need. \emph{Agent-friendly structure} gives agents task-relevant context, explicit interfaces, and bounded responsibilities for execution and revision.

Based on this observation, we introduce \textbf{RoboRSI}, a \textbf{multi-agent system} for robot self-improvement organized by \textbf{Top-Down Skill Refinement (TSR)}. TSR decomposes each task into compound, atomic, and base skills with scoped responsibilities and explicit input--output contracts, attributes each execution outcome to the responsible branch, and confines the revision to that branch: an inadequate decomposition revises the task structure, an unreliable composition the corresponding compound skill, and a failed operation the responsible atomic or base skill. A Manager, Planner, Engineer, and Reviewer carry out planning, execution, diagnosis, and validated release, and stable skill sequences are consolidated into compound skills that later executions select and monitor instead of reconstructing every tool decision.

We evaluate RoboRSI on a mobile manipulator that performs multi-object household cleanup across a scene change while its skills are developed over 104 rounds, and in simulation against robot-agent baselines on LIBERO, LIBERO-PRO, and RoboTwin, with frozen capability transfer to LIBERO-Plus. RoboRSI achieves the highest success rate on all four benchmarks, exceeding the strongest baseline by 5.3 points on LIBERO, 11.0 on LIBERO-PRO, 5.7 on LIBERO-Plus, and 2.7 on RoboTwin.

Our contributions are threefold:
\begin{itemize}
\item \textbf{Complementary abstractions for robot self-improvement:} we
identify that execution experience becomes reusable only when it is tied to
the task structure that gives it meaning, and that this structure must
serve both human-friendly steering and agent-friendly structure.
\item \textbf{RoboRSI Framework:} a system that realizes these abstractions through
Top-Down Skill Refinement, which verifies each atomic task's postcondition
before completion and revises each failure at its earliest failing node, and a multi-agent
execution--refinement loop that validates and releases skills across tasks.
\item \textbf{Real-world and simulated evaluation:} skill development on a
mobile manipulator over 104 rounds, the highest success
on four simulation benchmarks, and ablations of role separation, skill
refinement, and compound skills.
\end{itemize}

\section{Related Work}
\label{sec:related}
\label{sec:related-extended}
\paragraph{Code as Policy.}
Code as Policies~\citep{liang2023code} showed that language models can write
robot programs that compose perception and control APIs, turning high-level
instructions into executable behavior. Follow-up work examines what these
programs should be built from: CaP-X~\citep{fu2026capx} studies the
abstraction level of primitives, perceptual grounding, and interaction, and
GRACE~\citep{sun2025grace} encodes affordances and geometry as executable
concepts. ASPIRE~\citep{lu2026aspire} goes further by repairing failed
programs and keeping the repaired skills in a growing library. Once programs
are kept and reused, however, a repair to one skill can affect every task
that calls it, and deciding where a repair belongs becomes as important as
writing it. RoboRSI addresses this through TSR, which gives each capability
an interface and a declared responsibility, so that a repair is made at the
responsible branch and composes with the skills around it.

\paragraph{Recursive Self-Improvement (RSI).}
RSI studies how an agent's experience improves its later behavior. Language
agents first did so through verbal feedback: Reflexion stores reflections on
failed attempts, and Self-Refine iteratively critiques and revises its own
outputs~\citep{shinn2023reflexion,madaan2023selfrefine}. Voyager moved from
text to executable skills, accumulating a code library through open-ended
interaction~\citep{wang2023voyager}. Multimodal agents extend this idea with
skills acquired from external resources, execution feedback, and reusable
procedures, as in RESOURCE2SKILL, EvoSkill-GUI, and
OmniHarness~\citep{fan2026resource2skill,chen2026evoskillgui,xu2026omniharness}.
Robot-specific work detects and explains failures during execution:
ProgPrompt adds assertions to generated programs~\citep{singh2023progprompt},
REFLECT summarizes robot experience to explain and correct
failures~\citep{liu2023reflect}, DoReMi and Code-as-Monitor monitor
language- or code-specified constraints to trigger
recovery~\citep{guo2024doremi,zhou2025codeasmonitor}, and AHA trains a
vision-language model to detect and reason about manipulation
failures~\citep{duan2025aha}. These methods detect a failure within an
episode; RoboRSI additionally decides which skill should change and
validates the change before later tasks reuse it.
In embodied settings, EmbodiSkill diagnoses errors at the level of
individual skills~\citep{ju2026embodiskill}, and ENPIRE couples physical
evaluation with coding-agent refinement of policies~\citep{xiao2026enpire}.
Physical execution adds a difficulty that software environments largely
avoid: a failure may originate several steps before it is observed, and a
change to a shared skill affects tasks that were not part of the failed run.
RoboRSI therefore attributes each failure to its earliest failing node,
validates revisions against the recent history of the affected tasks, and
reuses capabilities through explicit skill interfaces.

\paragraph{Large Language Models for Robot.}
Large language models first entered robotics as planners. Early work
grounded generated plans in admissible actions~\citep{huang2022language},
and Inner Monologue fed environment feedback back into
planning~\citep{huang2022inner}. Advances in multimodal
understanding~\citep{alayrac2022flamingo,liu2023llava,gemini2024gemini15},
embodied grounding~\citep{driess2023palme}, and interleaved reasoning and
acting~\citep{yao2023react} then allowed models to reason directly over
observations, and VoxPoser~\citep{huang2023voxposer} and
ReKep~\citep{huang2024rekep} used generated code to compose 3D value maps and
relational keypoint constraints for motion planning. A parallel line trains
robot foundation models end to end. Vision--language--action models such as
RT-2, OpenVLA, $\pi_0$, and $\pi_{0.5}$ learn visuomotor control from large
robot datasets~\citep{brohan2023rt2,kim2024openvla,black2024pi0,physical2025pi05};
world-action models such as Zero-WAM add predictive visual
dynamics~\citep{zhou2026zerowam}; and Gemini Robotics and Gemini
Robotics-ER~1.5 combine control with embodied
reasoning~\citep{gemini2025robotics15}. As general-purpose models have
become more capable, recent systems use them directly for robot decisions
and actions. Maestro orchestrates perception and control
modules~\citep{shi2025maestro}, ETA/OpenETA closes the loop between planning,
execution, and feedback~\citep{chen2026eta}, GPT-Policy performs in-context
control~\citep{cheng2026gptpolicy}, two RoboHarness systems adapt through
memory~\citep{huang2026roboharness,li2026roboharness}, and UniDomain and
UniPlan connect symbolic planning with visual
grounding~\citep{ye2025unidomain,ye2026uniplan}. RoboRSI complements these
directions at the system level: TSR and multi-agent execution maintain
reusable skills, and the trajectories collected during execution can train
policies that return as tools in the same hierarchy.

\section{Method}
\label{sec:method}
Consider a coding agent connected directly to robot tools and asked to
improve a household cleanup task in a complex real-world home over many
runs. The context fills with
records of earlier runs; a failure observed late in a sequence is patched
where it appears, for example by adding retries to the release step,
although it originated in perception or navigation; and because the
objective is implicit in the latest prompt, successive edits optimize for
the latest scene. People are then drawn back into reading logs and deciding
where each fix belongs. RoboRSI removes this burden by making both the
intent of the people and the work of the agents explicit.

\subsection{Overall Framework}
\label{sec:framework}
RoboRSI turns repeated robot execution into a growing library of reusable
skills, organized around the two abstractions of Section~\ref{sec:intro}.
\textbf{Human-friendly steering} is the channel through which people enter
objectives, constraints, domain knowledge, and corrections at the level of
the task; each correction is kept as maintained guidance on the responsible
skill. \textbf{Agent-friendly structure} is the task--skill hierarchy in
which the agents work: every task and skill has an interface, a declared
responsibility, and a record of recent executions, so each agent receives
only the part of the hierarchy that its current step concerns.

\begin{figure}[!htbp]
\centering
\includegraphics[width=\linewidth]{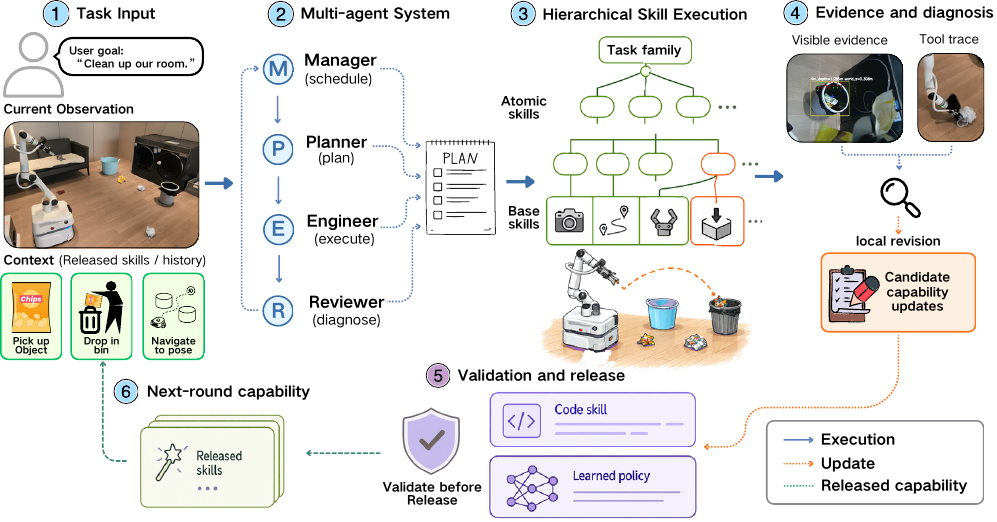}
\caption{\textbf{RoboRSI framework.} Given a task, the Manager organizes
its skill hierarchy, the Planner composes an execution plan, and the
Engineer runs the required skills. The Reviewer diagnoses failures from
observations and tool traces, returning them for TSR-guided revision.
Tested updates enter the next iteration, and stable branches become
parameterized skills for subsequent tasks.}
\label{fig:arch}
\end{figure}

\paragraph{Skill hierarchy.}
The hierarchy contains four kinds of nodes (Figures~\ref{fig:arch} and~\ref{fig:tsr-evolution}).
A \emph{task family}, such as household cleanup, groups tasks that share
goals and constraints. Each task is decomposed into \emph{atomic tasks},
scoped goals with an observable outcome such as placing an object in a
container, each achieved by an \emph{atomic skill} with a postcondition.
Atomic skills call \emph{base skills}, the perception and control
primitives exposed as tools. A stable sequence of skills is consolidated
into a \emph{compound skill}, a parameterized code policy invoked as a
single unit. Every skill declares its inputs, outputs, and responsibility,
and a skill may be shared by several branches.

Figure~\ref{fig:arch} follows one development round: the agents decompose
a stated goal and plan over released skills, execute them down to base
skills, locate the earliest failing node from visible evidence and the
tool trace, and validate and release the resulting updates for the next
round. People steer at the first stage and whenever a correction is needed.

\subsection{Top-Down Skill Refinement}
\label{sec:tsr}
\begin{figure}[!htbp]
\centering
\includegraphics[width=\linewidth]{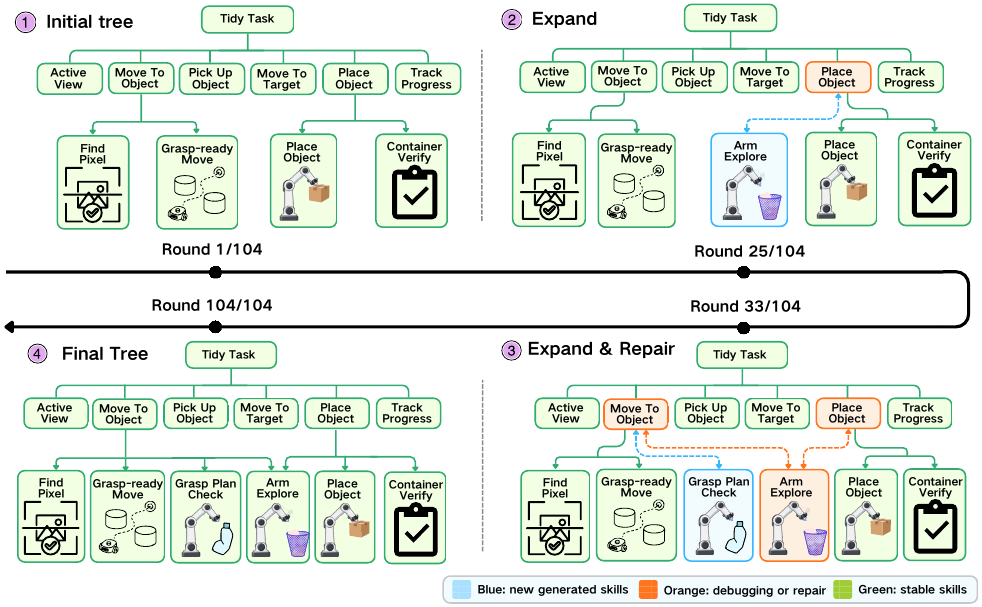}\\[1pt]
{\small (a) Task--skill structures during physical task development}\\[5pt]
\includegraphics[width=.9\linewidth]{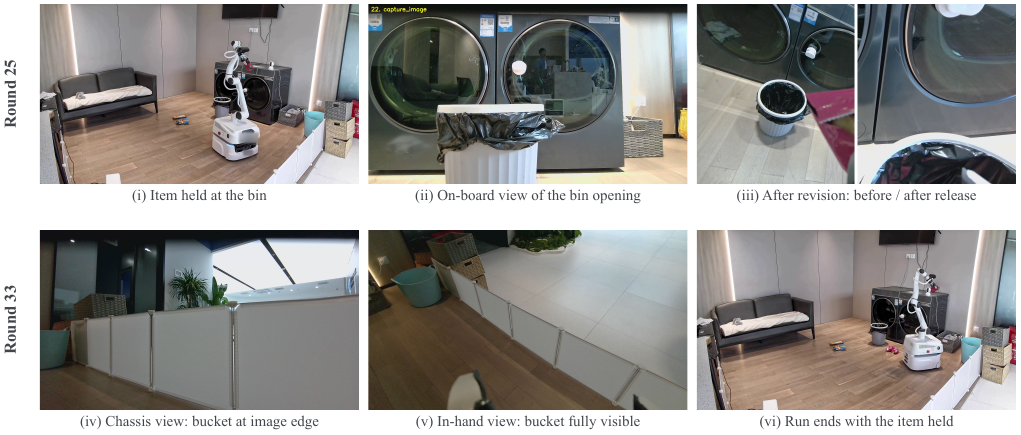}\\[1pt]
{\small (b) Two TSR revisions on the physical robot}
\caption{\textbf{TSR during physical cleanup.} (a) Four representative
task--skill structures show the progression from an initial task proposal
to a developed skill library; colored blocks mark the skill operations at
each snapshot. (b) Round~25 (top): the robot holds a wrapper at the correct
bin without releasing it (i), although the arm has reached its observation
pose (ii); the motion call reported a tolerance miss and the fallback
exhausted its budget. TSR traces the failure to this observation motion and
revises only the Place Object branch, which rechecks the reached pose and
executes the complete code-produced release sequence; the next placement
succeeds (iii). Round~33 (bottom): the destination bucket appears only at
the image edge in the chassis view (iv) but completely in the in-hand view
of the same sweep (v); the consumer rejected the producer's destination,
and the run ended with the second item held (vi). The revision leaves
identity to the producer and prefers complete views. Both revisions passed
offline replay of the four most recent runs and the existing regression
tests (Appendix~\ref{sec:physical-revisions}).}
\label{fig:tsr-evolution}
\end{figure}

Coding agents can solve local programming problems yet lose the task
objective across repeated edits, misinterpret physical conditions, or tailor
a fix too closely to one failed scene. TSR therefore organizes
implementation and revision around the skill hierarchy of
Section~\ref{sec:framework}: the task family and its atomic tasks preserve
the global objective, while atomic, compound, and base skills provide
bounded places for implementation, experience, and revision
(Figure~\ref{fig:tsr-evolution}a); Figure~\ref{fig:tsr-evolution}b
follows two revisions on the physical robot.

\paragraph{Design principles.}
Three principles connect local revision to reliable execution and reuse.
\emph{Responsibility and interface trust:} each skill reports its outcome
through an explicit interface, so callers rely on its postconditions
instead of duplicating its checks, and diagnosis follows the call path to
the responsible component instead of spreading fixes across callers.
\emph{Parameterization and history-guided revision:} object classes,
conditions, target relations, and execution order become parameters,
scene geometry comes from current observations, and each change is checked
against recent successful uses of the skill.
\emph{Persistent human knowledge:} a human correction becomes a runtime
check, a test, or maintained guidance on the responsible skill, so people
can concentrate on goals and exploration direction.

\paragraph{Refinement step.}
We write the released hierarchy at round $t$ as a graph $G_t=(V_t,E_t)$
whose nodes are tasks and skills and whose edges are possible calls; each
node $v$ carries an implementation or guidance $c_v$ and a declared
postcondition $\phi_v$. Executing a task produces a record $\tau_t$ of
observations and tool calls. We list the nodes executed on the task's branch
as $\pi_t=(v_1,\ldots,v_m)$ in the order in which they return, so that each
callee precedes its caller; a retried node appears once per attempt. The Reviewer evaluates the postconditions along this path and returns the earliest failing node
\begin{equation}
v_t^\star=v_{k^\star},\qquad
k^\star=\min\{k:\phi_{v_k}(\tau_t)=0\}.
\label{eq:tsr-attribution}
\end{equation}
The revision scope $S_t\subseteq V_t$ is the sub-hierarchy owned by
$v_t^\star$, and $\partial S_t$ denotes its interfaces to the rest of $G_t$.
A revision round is then
\begin{equation}
\Delta_t=\mathcal R\bigl(G_t|_{S_t},\partial S_t,\tau_t\bigr),\qquad
G_{t+1}=G_t\oplus\mathcal V(\Delta_t;\mathcal H_t).
\label{eq:tsr-refinement}
\end{equation}
Here $\mathcal R$ is the Reviewer's patch proposal, which may change only
nodes in $S_t$; if the interface itself must change, the scope is first
expanded to include its callers. $\mathcal H_t$ collects recent failed and
successful executions of every task whose call graph reaches $S_t$. The
validation operator $\mathcal V$ returns $\Delta_t$ if it passes review,
functional tests, and replay on $\mathcal H_t$ without regression, and the
empty patch otherwise; $\oplus$ applies the accepted patch.
Eq.~(\ref{eq:tsr-attribution}) places each repair at the origin of the
failure rather than where it is observed, the restriction to $S_t$ bounds
its effect to the tasks that reach $S_t$, and validation on $\mathcal H_t$
preserves behavior that already works (Appendix~\ref{sec:tsr-analysis}).

\paragraph{From stable branches to reusable capabilities.}
Stable branches become compound skills, shifting online reasoning from
individual tool choices to skill selection and monitoring; execution
trajectories can also train neural policies exposed as skills in the same
hierarchy (Section~\ref{sec:policy-learning-study}).

\subsection{Multi-Agent System}
\label{sec:roles}
Four agents carry out each round within this structure. The
\textbf{Manager} decomposes the goal into atomic tasks, identifies the
base skills each one needs, maintains skill versions, reviews each proposed
revision, may request or author changes to it, and decides whether it is
admitted, which branch it may change, and whether it is released.
The \textbf{Planner} receives one atomic task, the current observation, and
the interfaces of the skills below it, and returns an executable plan.
The \textbf{Engineer} executes the plan through the tool interface and
implements missing skills.
The \textbf{Reviewer} judges from the observations and tool trace whether
an atomic task reached its expected outcome and, for a failure, identifies
where execution first diverged, and writes a revision patch for the
responsible branch that it submits to the Manager.
Each agent works from the part of the hierarchy that its step concerns, so
its context stays short and every change can be traced to a diagnosis.

\section{Experiment}
\label{sec:experiments}
We report physical deployment and simulation results, then compare
individual episodes and summarize the ablations.

\subsection{Physical Robot Case Study}
\label{sec:real-world}

\paragraph{Setting.}
We study household floor cleanup with a WheelSingleArm M1 mobile
manipulator. The robot searches for objects, approaches and grasps them,
and transports them to designated receptacles. The case study covers
104 physical development runs with 24 hours of cumulative operation. Before each run, the scene is restored
to a common base layout, while the objects and receptacles are partially
randomized; the navigation map remains fixed. Platform details and the
reset procedure are given in Appendix~\ref{sec:mobile-real-method}.

\begin{figure}[!htbp]
    \centering
    \includegraphics[width=0.7\linewidth]{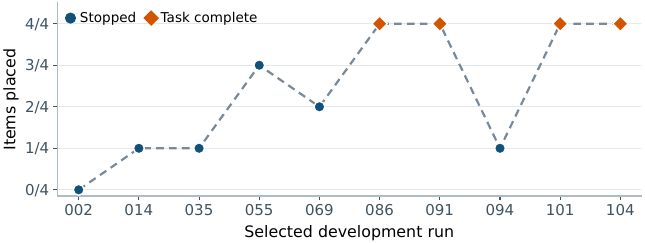}
    \caption{\textbf{Placement progress in selected physical cleanup runs.}
    Each point shows the number of objects placed among the first four
    items considered in that run; placements were confirmed on site.
    Marker colors indicate the audited task outcome. Points are equally
    spaced in selected-run order, while tick labels give the development
    run numbers. Dashed segments connect selected observations only.
    The four-item denominator is a common display reference: task
    manifests and object configurations vary across runs.}
    \label{fig:physical-four-item-progress}
\end{figure}

\paragraph{Iterative cleanup.}
Figure~\ref{fig:physical-four-item-progress} follows ten selected
runs from the 104-run development period. In the earliest selected
run, the robot placed no objects. Later runs progressed to placing
several objects, and some completed their assigned cleanup tasks
and returned to the dock. The change was not steady: a later run
recorded only one placement after complete runs had already
appeared. We counted an object only when on-site personnel observed
it enter its designated receptacle. The figure displays at most
four placements per run, although the task manifests varied in size.

The skill-tree snapshots in Figure~\ref{fig:tsr-evolution}a provide a
complementary view of this development process. The hierarchy introduces
\emph{Arm Explore} under \emph{Place Object} to obtain a view of the target
receptacle. Later revisions add \emph{Grasp Plan Check} under
\emph{Move To Object} and make \emph{Arm Explore} available to both the
approach and placement branches. These changes target two recurring
requirements of the cleanup procedure: checking a candidate grasp
before motion and grounding placement in an observation of the
destination. The snapshots document how the corresponding capabilities
were organized and reused; the selected executions do not isolate the
effect of an individual revision.

\vspace{-0.2cm}

\begin{figure}[!htbp]
    \centering
    \includegraphics[width=\linewidth]{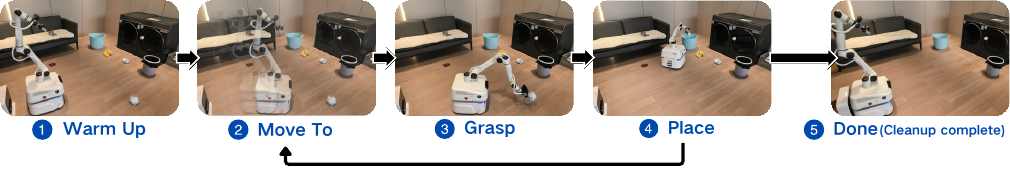}
    \caption{\textbf{Two-scene physical mobile cleanup.}
    Selected frames show preparation, approach, grasping, placement,
    and the recorded final scene. The return arrow indicates repeated
    object handling. Objects and receptacles were rearranged between
    scenes, while the agent configuration remained unchanged.}
    \label{fig:real-world}
\end{figure}

\paragraph{Physical execution.}
The two-scene demonstration provides an extended execution example.
The robot placed all five designated objects in the first scene and
all four in the second, with each placement confirmed on site. Between
scenes, an operator rearranged the objects and receptacles; the skill
code, prompts, and policy configuration were not modified. The cleanup
executions required no manual correction or restart. Following the
scene change, the robot repeated the search, approach, grasp, transport,
and placement sequence. Figure~\ref{fig:real-world} illustrates stages
of this recorded execution.

\subsection{Evaluation Design}
\label{sec:eval-design}
\paragraph{Choice of baselines.}
RoboRSI accumulates skills serially: each episode runs with the current
library, and revisions that pass validation are released to the episodes
that follow. To our knowledge, no existing robot-agent system accumulates
skills in this serial form. ASPIRE~\citep{lu2026aspire}, the closest
self-improving code agent, learns every task in parallel with a dedicated
agent that runs a population-based search over candidate programs, admits
the resulting skills after the search, and evaluates one generated program
per task; ENPIRE~\citep{xiao2026enpire} refines learned policies and their
training code through physical experimentation. Both rely on development
loops that differ from a single serial stream, so they cannot be run under
the same protocol. We therefore compare with the strongest agents that act
through code or tools, CaP-X, Maestro, and OpenETA, with the same backbone.

\paragraph{Self-improvement as the evaluated capability.}
All methods are evaluated under the same conditions: the same backbone,
tool interface, per-episode interaction budget, task panels, and simulator
verdict. The baselines execute with a fixed set of tools and skills and do
not revise or add skills across episodes; RoboRSI differs only in that it
can use the outcome of an episode to revise its skills for later episodes. Improving on the tasks at hand is
the capability that RoboRSI contributes, so the comparison on LIBERO,
LIBERO-PRO, and RoboTwin measures what self-improvement adds over strong
fixed agents. LIBERO-Plus complements this by evaluating the library with
self-improvement disabled on perturbed instances.

\paragraph{Verification before completion.}
Checking the postcondition of every atomic task against fresh observations
before the agent proceeds is part of RoboRSI's design (Section~\ref{sec:tsr}
and Appendix~\ref{sec:tsr-analysis}), not an external addition. The lower
share of premature completions in Figure~\ref{fig:failure-analysis}a is one
measured effect of this design.

\paragraph{Evidence for the components.}
Section~\ref{sec:ablation} isolates the components of the system: role
separation (36 versus 9 of 50 RoboTwin tasks), TSR against flat revision on
the same failure, compound skills over 600 matched episode pairs ($+7.5$
points, McNemar $p<10^{-4}$), the backbone model, and policies learned from
execution data.

\paragraph{Scale of the evaluation.}
The main comparison comprises 5,974 evaluation episodes: 150 per method on
LIBERO and 600 on LIBERO-PRO, each with five initial states per task, 840
perturbation instances per method on LIBERO-Plus, and 150 episodes per
baseline and 154 for RoboRSI on RoboTwin. The ablations add 2,770 episodes,
and the physical study covers 104 development rounds. In total, RoboRSI is
evaluated on 230 tasks across four benchmarks with 1,744 episodes. For reference, the main
LIBERO evaluation of OpenETA uses 40 tasks with ten seeds each (400
episodes), and its self-evolution studies use at most ten
tasks~\citep{chen2026eta}; ASPIRE evaluates 69 tasks on LIBERO-PRO,
Robosuite, and BEHAVIOR-1K~\citep{lu2026aspire}; the memory-augmented harness of
\citet{li2026roboharness} is evaluated on 25 tasks, and
\citet{huang2026roboharness} on LIBERO, LIBERO-Plus, and five long-horizon
tasks.

\paragraph{Implementation details.}
The postcondition of each skill, the release gate, the construction of the
replay history, and the consolidation of compound skills are specified in
the descriptor of each skill and summarized in
Appendix~\ref{sec:tsr-analysis}.

\subsection{Main Simulation Results}
\label{sec:simulation}
\paragraph{Evaluation settings.}
Self-improvement is part of the evaluated capability: on LIBERO,
LIBERO-PRO, and RoboTwin, RoboRSI starts from its initial skill library and
improves online on the evaluation tasks, summarizing failures and revising
skills as evaluation proceeds, whereas the baselines have no
self-improvement mechanism (Section~\ref{sec:eval-design}). LIBERO-Plus instead tests transfer: the library
obtained on LIBERO is held fixed and receives no updates on the perturbed
instances. All methods use GPT-5.6-SOL. Maestro, the most costly baseline,
is evaluated on LIBERO and LIBERO-PRO only.

\begin{table}[!htbp]
\centering\small
\caption{\textbf{Simulation results.} Success is successful/valid episodes (rate);
coverage is solved/total tasks (rate), counting each task once after its
first success. LIBERO uses 30 tasks$\times$5 episodes, LIBERO-PRO uses
120 tasks$\times$5 episodes, LIBERO-Plus uses 840 perturbation instances of 30 tasks,
and RoboTwin uses 50 tasks$\times$3 episodes for the baselines; for RoboRSI we
report the first 154 episodes of its online run on the same 50 tasks. RoboRSI improves online on
LIBERO, LIBERO-PRO, and RoboTwin and uses its fixed LIBERO library on
LIBERO-Plus. Maestro is evaluated on LIBERO and LIBERO-PRO because of its
computational cost.}
\label{tab:main-simulation}
\setlength{\tabcolsep}{9pt}
\renewcommand{\arraystretch}{1.0}
\begin{tabular}{llrr}
\toprule
Benchmark & Method & Success & Coverage \\
\midrule
LIBERO & CaP-X & 36/150 (24.0\%) & 16/30 (53.3\%) \\
 & Maestro & 62/150 (41.3\%) & 23/30 (76.7\%) \\
 & OpenETA & 76/150 (50.7\%) & 23/30 (76.7\%) \\
\rowcolor{rsirow} & \textbf{RoboRSI} & \textbf{84/150 (56.0\%)} & \textbf{24/30 (80.0\%)} \\
\midrule
LIBERO-PRO & CaP-X & 111/600 (18.5\%) & 46/120 (38.3\%) \\
 & Maestro & 198/600 (33.0\%) & 77/120 (64.2\%) \\
 & OpenETA & 231/600 (38.5\%) & 80/120 (66.7\%) \\
\rowcolor{rsirow} & \textbf{RoboRSI} & \textbf{297/600 (49.5\%)} & \textbf{102/120 (85.0\%)} \\
\midrule
LIBERO-Plus & CaP-X & 130/840 (15.5\%) & 21/30 (70.0\%) \\
 & OpenETA & 306/840 (36.4\%) & 27/30 (90.0\%) \\
\rowcolor{rsirow} & \textbf{RoboRSI} & \textbf{354/840 (42.1\%)} & \textbf{28/30 (93.3\%)} \\
\midrule
RoboTwin & CaP-X & 3/150 (2.0\%) & 3/50 (6.0\%) \\
 & OpenETA & 32/150 (21.3\%) & 17/50 (34.0\%) \\
\rowcolor{rsirow} & \textbf{RoboRSI} & \textbf{37/154 (24.0\%)} & \textbf{26/50 (52.0\%)} \\
\bottomrule
\end{tabular}
\end{table}

\paragraph{Task execution and coverage.}
RoboRSI achieves the highest success and coverage on every benchmark
(Table~\ref{tab:main-simulation}). The largest advantage is on LIBERO-PRO,
where it exceeds OpenETA by 11.0 points in success and 18.3 points in
coverage, so the gain extends to a larger set of solved tasks. On LIBERO, it
exceeds OpenETA by 14.0 points on Spatial and 8.0 points on Object
(Section~\ref{sec:libero-suites}).

\paragraph{Where the LIBERO-PRO gains come from.}
RoboRSI is best in every suite and under every perturbation
(Table~\ref{tab:pro-breakdown}). The margins are largest on the Object suite
($+16.5$) and under object perturbations ($+15.3$), which require
re-identifying the target in the current scene.

\begin{table}[!htbp]
\centering\small
\caption{\textbf{LIBERO-PRO by task suite and perturbation type.} Success
rate (\%); each suite contains 200 episodes and each perturbation type 150
episodes per method. Bold: best in the column; gray: RoboRSI's gain over the
strongest baseline in percentage points.}
\label{tab:pro-breakdown}
\setlength{\tabcolsep}{3.6pt}
\renewcommand{\arraystretch}{1.12}
\begin{tabular}{lcccccccc}
\toprule
 & \multicolumn{3}{c}{Task suite} & \multicolumn{4}{c}{Perturbation type} & \\
\cmidrule(lr){2-4}\cmidrule(lr){5-8}
Method & Spatial & Object & Goal & Language & Object & Position & Task & All \\
\midrule
CaP-X & 24.5 & 13.5 & 17.5 & 21.3 & 16.7 & 17.3 & 18.7 & 18.5 \\
Maestro & 39.5 & 38.0 & 21.5 & 34.7 & 30.0 & 32.0 & 35.3 & 33.0 \\
OpenETA & 50.0 & 36.5 & 29.0 & 42.7 & 33.3 & 36.0 & 42.0 & 38.5 \\
\rowcolor{rsirow}RoboRSI & \textbf{57.0}\gain{+7.0} & \textbf{54.5}\gain{+16.5} & \textbf{37.0}\gain{+8.0} & \textbf{52.0}\gain{+9.3} & \textbf{48.7}\gain{+15.3} & \textbf{45.3}\gain{+9.3} & \textbf{52.0}\gain{+10.0} & \textbf{49.5}\gain{+11.0} \\
\bottomrule
\end{tabular}

\end{table}

\begin{figure}[!htbp]
\centering
\includegraphics[width=\linewidth]{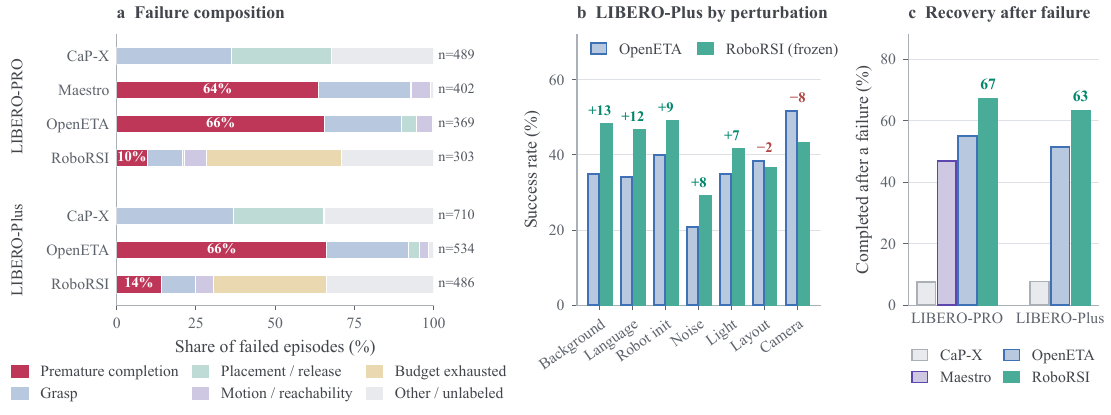}
\caption{\textbf{Failure analysis.} (a) Composition of failed episodes on
LIBERO-PRO and LIBERO-Plus; $n$ is the number of failed episodes, and the
premature-completion share is labeled. (b) Success on LIBERO-Plus by
perturbation type, 120 instances each; numbers give RoboRSI's difference
from OpenETA in percentage points. (c) Share of episodes that still succeed
after an execution failure, over the 209 LIBERO-PRO and 258 LIBERO-Plus
(task, seed) pairs in which every method has a complete trace and at least
one execution failure (Appendix~\ref{sec:failure-taxonomy}).}
\label{fig:failure-analysis}
\end{figure}

\paragraph{Transfer under perturbations.}
On LIBERO-Plus, the frozen source library exceeds OpenETA by 5.7 points and
solves at least one perturbed instance of 28 of the 30 tasks. The gains are
largest under background ($+13$), language ($+12$), robot initial state
($+9$), noise ($+8$), and lighting ($+7$) perturbations
(Figure~\ref{fig:failure-analysis}b reports all seven types).

\paragraph{Why the baselines fail.}
About two thirds of the failures of OpenETA and Maestro on LIBERO-PRO (66\%
and 64\%) and of OpenETA on LIBERO-Plus (66\%) are premature completions:
the agent declares the task done after a tool reports a release, while the
simulator judges it unfinished (Figure~\ref{fig:failure-analysis}a). For
RoboRSI, premature completion accounts for 10\% and 14\% of failures. The
methods differ in how they decide that a task is complete: the baselines
accept a successful tool return, whereas every RoboRSI atomic task has a
postcondition checked against fresh observations before the agent proceeds
(Eq.~(\ref{eq:tsr-attribution})). The failure composition reflects this
completion rule, and RoboRSI's remaining failures mostly end with the budget
exhausted during recovery rather than a false claim of success. On episodes
with an execution failure, it still completes 67.0\% on
LIBERO-PRO and 63.2\% on LIBERO-Plus, against 55.0\% and 51.6\% for OpenETA
(Figure~\ref{fig:failure-analysis}c).

\subsubsection{LIBERO Results by Suite}
\label{sec:libero-suites}
Table~\ref{tab:libero-suites} splits the LIBERO results of
Table~\ref{tab:main-simulation} by suite; each suite has 10 tasks and 50
episodes per method.

\begin{table}[!htbp]
\centering\small
\caption{\textbf{LIBERO by suite.} Successful/valid episodes per suite.}
\label{tab:libero-suites}
\begin{tabular}{lcccc}
\toprule
Method & Spatial & Object & Goal & All \\
\midrule
CaP-X & 12/50 & 15/50 & 9/50 & 36/150 \\
Maestro & 22/50 & 24/50 & 16/50 & 62/150 \\
OpenETA & 23/50 & 29/50 & \textbf{24/50} & 76/150 \\
\rowcolor{rsirow}\textbf{RoboRSI} & \textbf{30/50} & \textbf{33/50} & 21/50 & \textbf{84/150} \\
\bottomrule
\end{tabular}
\end{table}

\subsubsection{One-Day Self-Iteration}
\label{sec:one-day}
\begin{figure}[!htbp]
\centering
\includegraphics[width=.55\linewidth]{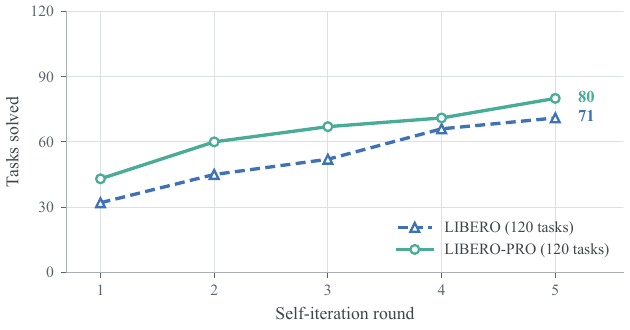}
\caption{\textbf{One-day self-iteration.} Cumulative number of tasks solved
at least once after each self-iteration round; each task is counted at its
first recorded success.}
\label{fig:evolution-cost}
\end{figure}

To examine self-iteration without human steering, we let RoboRSI iterate on
its own for one day on a 120-task LIBERO catalog and on the 120 LIBERO-PRO
tasks. In each round, the current library runs on the tasks not yet solved,
and revisions that pass validation are released for the next round.
Figure~\ref{fig:evolution-cost} reports the cumulative number of tasks
solved at least once. Within five rounds, coverage grows from 32 to 71 of
120 LIBERO tasks and from 43 to 80 of 120 LIBERO-PRO tasks; the five LIBERO
rounds take 7.0 hours of active execution. These counts accumulate across
released library versions and are separate from the evaluation in
Table~\ref{tab:main-simulation}.

\subsection{Case Study}
\label{sec:skill-development}
\label{sec:sim-cases}
\begin{figure}[!htbp]
\centering
\includegraphics[width=.7\linewidth]{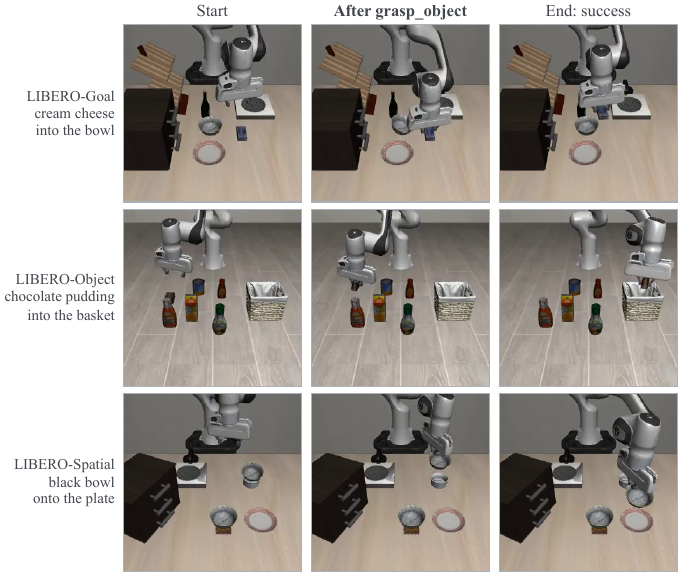}
\caption{\textbf{Reuse of a revised skill on new tasks.} The revised
\texttt{grasp\_object}, developed on a single task, is invoked on three
tasks outside its development set, each of which had failed in the earlier
frozen evaluation. Columns show the first recorded frame, the state after
the \texttt{grasp\_object} call, and the final successful state.}
\label{fig:skill-reuse}
\end{figure}

\begin{figure}[!htbp]
\centering
\includegraphics[width=\linewidth]{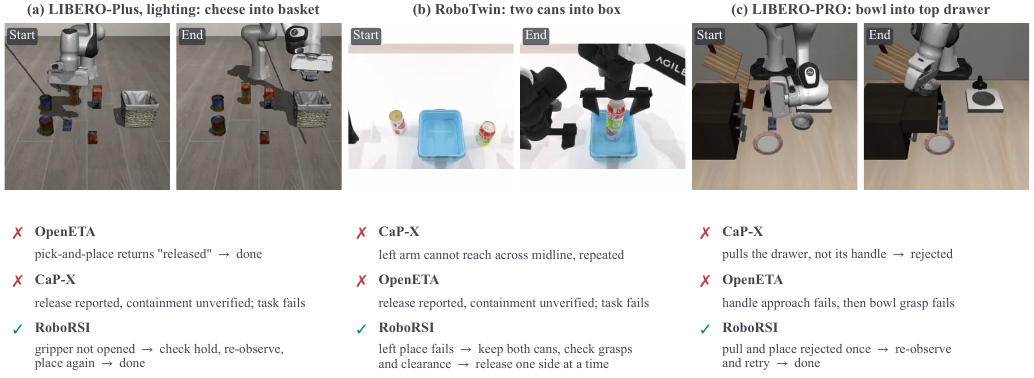}
\caption{\textbf{Simulation cases with the same initial state across
methods.} Images show the first and last recorded frames of the RoboRSI
episode; text summarizes the decisive tool calls of each method.}
\label{fig:sim-cases}
\end{figure}

We examine the three episodes of Figure~\ref{fig:sim-cases}, in which
RoboRSI succeeds while the baselines fail from the same initial state, and
then follow a
revised skill as it is reused on tasks it was not developed on
(Figure~\ref{fig:skill-reuse}).

\paragraph{Terminal checks under a lighting perturbation.}
On LIBERO-Plus, the task is to place a cream-cheese package in a basket.
OpenETA's combined pick-and-place tool returned that the object had been
released at the destination, and the agent declared success at once. CaP-X's
placement also reported a release while its containment check remained
unverified. Both episodes failed. RoboRSI, with its frozen source library,
received a placement result stating that the gripper had not opened. Rather
than ending, it checked whether it still held the package, re-observed the
basket from the head and top views, placed the package again, and succeeded.

\paragraph{Dual-arm state after a failed placement.}
On RoboTwin, both cans must be picked up with separate arms and placed into
a plastic box. CaP-X repeatedly tried to reach the box with the left arm
across the midline, which was infeasible. OpenETA stopped after the gripper
reported a release, with containment unverified, and the task failed. When
RoboRSI's first left-arm placement failed, it kept both cans in hand,
verified both grasps and the clearance between the loaded arms, and then
released the cans one side at a time.

\paragraph{Recovery in a multi-step task.}
On LIBERO-PRO, the robot must open the top drawer and put a bowl inside.
CaP-X invoked the pulling skill on the drawer rather than its handle and
was rejected; OpenETA repeatedly failed to approach the handle and then
failed to grasp the bowl. RoboRSI also had a pull and a placement rejected,
but continued to observe and act until the task succeeded.

In all three episodes, the baselines fail by trusting a tool's return
value, whereas RoboRSI's skills expose failed outcomes and its agents verify
postconditions before proceeding, the mechanism behind the failure
composition in Figure~\ref{fig:failure-analysis}a.

\paragraph{Reusing a revised skill across tasks.}
Revisions made during development are released as shared skills and reach
tasks they were never developed on. The base skill \texttt{grasp\_object} was
revised on a single development task: the Manager's patch makes every retry
verify the target's identity from a fresh observation and forbids replaying
a grasp point that has already failed, and the revision passed its gate
before release. After release, 42 successful episodes on other tasks invoke
the revised skill. Figure~\ref{fig:skill-reuse} shows three of them, from the
Goal, Object, and Spatial suites, each of which had failed in the earlier
frozen evaluation. Because the skill is shared through the hierarchy, a
repair made in one task becomes available to every branch that calls it.

\subsection{Ablation Study}
\label{sec:ablation}
\label{sec:ablation-main}
We study role separation, hierarchical refinement, code consolidation, and
backbone choice separately, and additionally examine policy learning from
execution data.
The backbone study uses LIBERO-PRO and RoboTwin task panels with three evaluation
episodes per task, 360 and 150 episodes per backbone.

\subsubsection{Role Separation}
\label{sec:single-multi}
We compare development on the 50 RoboTwin tasks with a single agent and with
separate roles (Table~\ref{tab:single-multi}). In the single-agent condition,
one Engineer plans, executes, and judges its own attempts. In the
multi-agent condition, a Planner composes the plan, the Engineer executes
it, and a Reviewer diagnoses each outcome from observations and tool traces.
The single agent solves 9 of the 50 tasks, whereas the multi-agent system
solves 36. Separating diagnosis from execution gives every failure an
independent review before the next attempt, instead of leaving the
executing agent to judge its own result.

\begin{table}[!htbp]
\centering\small
\caption{\textbf{Single-agent and multi-agent development on RoboTwin.}
Tasks solved at least once over the full online run, out of 50; the
multi-agent condition is the RoboTwin run of Table~\ref{tab:main-simulation},
whose first 154 episodes are reported there.}
\label{tab:single-multi}
\setlength{\tabcolsep}{8pt}
\renewcommand{\arraystretch}{1.12}
\begin{tabular}{llc}
\toprule
Condition & Roles & Tasks solved \\
\midrule
Single agent & Engineer & 9/50 (18.0\%) \\
\rowcolor{rsirow}Multi-agent & Planner, Engineer, Reviewer & \textbf{36/50 (72.0\%)} \\
\bottomrule
\end{tabular}
\end{table}

\subsubsection{Top-Down Skill Refinement}
\label{sec:ablation-tsr}
TSR determines where the system improves. We therefore compare TSR with
flat revision on how well each turns an observed failure into a validated,
released improvement, rather than on task success. Both conditions
start from the same code and memory and receive the same public traces of a
failed \texttt{libero\_pick\_place} execution on \texttt{libero\_goal/8}.
They may modify Atomic and Base skills, and use the same Planner, Engineer,
Reviewer, and Manager review. A candidate is released only if the Base
functional checks pass and the complete Atomic task succeeds on both
development seeds. Under TSR, the Reviewer first builds a responsibility tree
from the Atomic failure and selects the layers to revise; under flat revision,
it proposes a revision directly from the failure.

TSR attributed the failure to two nodes: the Atomic plan treated a
do-not-regrasp flag set at pickup as permanent after a verified release, and
the Base hold check used the fixed pickup patch as an invariant during
transport. Its candidate changes only these nodes, succeeds on both seeds,
and is released (Table~\ref{tab:ablation-tsr}). The flat candidate also
rewrites the placement policy and succeeds on one seed; on the other, the
agent declares success with the object off the target. A further flat
candidate, restricted to the Atomic plan after Manager review, again
succeeds on one seed, so no flat candidate is released.

\begin{table}[!htbp]
\centering\small
\caption{\textbf{TSR and flat revision on the same failed task.} Files and
lines count changes relative to the common starting code; Atomic success is
over the two development seeds.}
\label{tab:ablation-tsr}
\begin{tabular}{lcccc}
\toprule
Condition & Files & Changed lines & Atomic success & Released \\
\midrule
\rowcolor{rsirow}\textbf{TSR} & 3 & 50 & \textbf{2/2} & yes \\
Flat, candidate 1 & 4 & 104 & 1/2 & no \\
Flat, candidate 2 & 1 & 63 & 1/2 & no \\
\bottomrule
\end{tabular}
\end{table}

\subsubsection{Code Consolidation}
\label{sec:ablation-code}
\begin{figure}[!htbp]
\centering
\includegraphics[width=.8\linewidth]{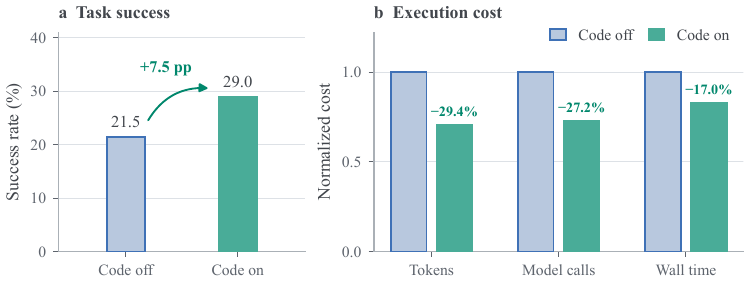}
\caption{\textbf{Effect of code consolidation.} (a) Task success over 600 matched
episodes per condition. (b) Median token, model-call, and time cost on the 118-task
efficiency sample, normalized to Code off.}
\label{fig:ablation-code}
\end{figure}

We compare execution with compound skills available to the agent
(Code on) and withheld (Code off) on a separate panel of 120 tasks selected
from LIBERO-Plus, distinct from the task sets in Table~\ref{tab:main-simulation},
so its absolute success rates are not comparable with the main results. The
base tools, the GPT-5.6-SOL backbone, the 120 tasks, and the five seeds are
the same in both conditions, so every Code-on episode has a matched Code-off
counterpart.
Figure~\ref{fig:ablation-code} reports success over these 600 matched pairs
and normalized median costs on a separate 118-task sample.

Code consolidation raises success from 21.5\% (129/600) to 29.0\% (174/600),
a paired gain of 7.5 percentage points (95\% CI $[3.7, 11.5]$; McNemar
$p<10^{-4}$), and increases the number of tasks solved at least once from 58
to 71 of 120. The gain is spread across many episodes: 86 matched pairs
succeed only with Code on, against 41 that succeed only with Code off. Median token consumption, model calls, and wall time decrease by
29.4\%, 27.2\%, and 17.0\%, respectively.

The success gain follows from how errors arise in long tool sequences.
Without code, the agent reconstructs the sequence in every episode, and
each step is another opportunity to choose the wrong tool, pass an
inconsistent argument, or skip a check between steps; over sequences of
many steps, these errors accumulate. A compound skill fixes the validated
order of calls and the checks between them, so the agent decides only which
skill to invoke and with which parameters. The physical Round~25 failure in
Section~\ref{sec:tsr} shows the same effect: a model-written release
sequence dropped steps that the code-produced sequence contained. The cost
reductions have the same source, since one skill call replaces several
model decisions. Wall time falls less than tokens and model calls because
the motion executed in the simulator is unchanged; only the reasoning
between motions is removed. The agent still responds to execution feedback
and returns to step-by-step reasoning when a compound skill reports a
failure.

\subsubsection{Backbone Model}
\label{sec:ablation-model}
We compare GPT-5.6-SOL, GPT-5.5, and GPT-5.4 across the reasoning roles with
the same initial skills, task distribution, and interaction budget
(Table~\ref{tab:ablation-model}). GPT-5.5 reaches the highest LIBERO-PRO
success, 48.6\%, and leads on the Object and Goal suites, while GPT-5.6-SOL
is best on RoboTwin. The number of model calls hardly differs, with medians
of 115 to 124 per LIBERO-PRO episode, so the stronger backbone does not win
by reasoning longer. The difference lies in how often an episode runs out of
budget: GPT-5.4 exhausts it in 101 episodes and GPT-5.5 in 80, so the
stronger backbone reaches a verified outcome within the budget more often.
Because the ranking changes between benchmarks while the task structure and
skill library stay the same, the choice of backbone interacts with the
benchmark rather than favoring one model throughout.

\begin{table}[!htbp]
\centering\small
\caption{\textbf{Backbone comparison.} The study uses a separate panel of
120 LIBERO-PRO tasks$\times$3 seeds run independently of
Table~\ref{tab:main-simulation}, so only the relative differences between
backbones are compared. LIBERO-PRO success over 360 episodes per backbone,
overall and by task suite; RoboTwin successes over 150
episodes. Model calls are medians per LIBERO-PRO episode; budget exhausted
counts LIBERO-PRO episodes that ended at the interaction limit. Bold: best in
the column.}
\label{tab:ablation-model}
\setlength{\tabcolsep}{4.5pt}
\renewcommand{\arraystretch}{1.12}
\begin{tabular}{lccccccc}
\toprule
 & \multicolumn{4}{c}{LIBERO-PRO success (\%)} & RoboTwin & \multicolumn{2}{c}{LIBERO-PRO episodes} \\
\cmidrule(lr){2-5}\cmidrule(lr){7-8}
Backbone & All & Spatial & Object & Goal & success & Model calls & Budget exhausted \\
\midrule
GPT-5.6-SOL & 38.3 & 38.3 & 48.3 & 28.3 & \textbf{24/150} & 115 & 79 \\
GPT-5.5 & \textbf{48.6} & 53.3 & \textbf{55.0} & \textbf{37.5} & 19/150 & 122 & 80 \\
GPT-5.4 & 40.3 & \textbf{55.8} & 39.2 & 25.8 & 21/150 & 124 & 101 \\
\bottomrule
\end{tabular}

\end{table}

\subsubsection{Learning Policies from Execution Data}
\label{sec:policy-learning-study}
\begin{figure}[!htbp]
\centering
\includegraphics[width=.72\linewidth]{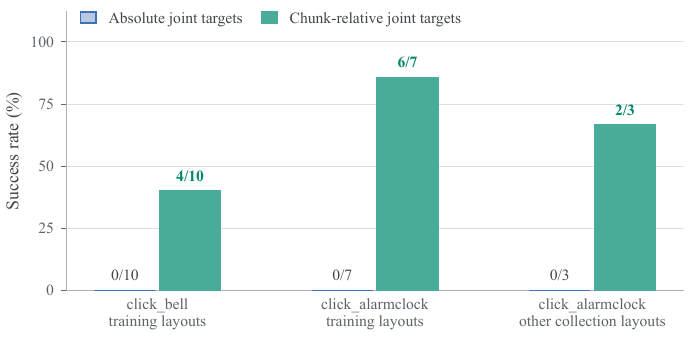}
\caption{\textbf{ACT policies trained on RoboRSI executions.} Success with
absolute and chunk-relative joint targets under the same demonstrations,
architecture, and training budget.}
\label{fig:act-results}
\end{figure}

Successful RoboRSI executions provide demonstrations for training compact
visuomotor policies, which can then be exposed as skills in the hierarchy.
We train ACT policies~\citep{zhao2023act} on 10 successful \texttt{click\_bell} and 7 successful
\texttt{click\_alarmclock} executions and compare two action representations
with the same data, architecture, and 10,000 training steps. With absolute
joint targets, neither task is solved in any evaluated layout. Expressing
the joint targets relative to the arm configuration at the start of each
action chunk raises success to 4/10 on the bell training layouts, and on
the alarm clock to 6/7 on training layouts and 2/3 on collection layouts
that did not supply training data (Figure~\ref{fig:act-results}).
With so few demonstrations, an absolute target requires the policy to
regress the full arm configuration at every step, and small regression
errors move the end effector away from the small contact region that these
pressing tasks require. A chunk-relative target encodes only how the arm
moves from its current configuration, which the demonstrations constrain
far more tightly: on the training observations, the joint-command error
falls by 57\%, from 0.0086 to 0.0037\,rad.
Appendix~\ref{sec:learned-policy-experiments} gives the training details.

\section{Conclusions and Future Work}
\label{sec:conclusion}
RoboRSI combines Top-Down Skill Refinement and multi-agent coordination to
organize execution, diagnosis, local revision, and reusable capability,
toward stable, efficient, and reusable robot self-evolution in complex
real-world environments. Future work
will transfer simulation-acquired experience to physical robots and share
skills across robots and platforms.

\clearpage
\ULforem
\bibliography{refs}

@inproceedings{liang2023code,
  title={Code as Policies: Language Model Programs for Embodied Control},
  author={Liang, Jacky and Huang, Wenlong and Xia, Fei and Xu, Peng and Hausman, Karol and Ichter, Brian and Florence, Pete and Zeng, Andy},
  booktitle={IEEE International Conference on Robotics and Automation (ICRA)},
  year={2023},
  url={https://arxiv.org/abs/2209.07753}
}

@inproceedings{yao2023react,
  title={{ReAct}: Synergizing Reasoning and Acting in Language Models},
  author={Yao, Shunyu and Zhao, Jeffrey and Yu, Dian and Du, Nan and Shafran, Izhak and Narasimhan, Karthik and Cao, Yuan},
  booktitle={International Conference on Learning Representations},
  year={2023},
  url={https://arxiv.org/abs/2210.03629}
}

@inproceedings{huang2022inner,
  title={Inner Monologue: Embodied Reasoning through Planning with Language Models},
  author={Huang, Wenlong and Xia, Fei and Xiao, Ted and others},
  booktitle={Conference on Robot Learning},
  year={2022},
  url={https://arxiv.org/abs/2207.05608}
}

@inproceedings{shinn2023reflexion,
  title={Reflexion: Language Agents with Verbal Reinforcement Learning},
  author={Shinn, Noah and Cassano, Federico and Gopinath, Ashwin and Narasimhan, Karthik and Yao, Shunyu},
  booktitle={Advances in Neural Information Processing Systems},
  year={2023},
  url={https://arxiv.org/abs/2303.11366}
}

@article{wang2023voyager,
  title={Voyager: An Open-Ended Embodied Agent with Large Language Models},
  author={Wang, Guanzhi and Xie, Yuqi and Jiang, Yunfan and Mandlekar, Ajay and Xiao, Chaowei and Zhu, Yuke and Fan, Linxi and Anandkumar, Anima},
  journal={Transactions on Machine Learning Research},
  year={2024},
  url={https://arxiv.org/abs/2305.16291}
}

@inproceedings{kim2024openvla,
  title={{OpenVLA}: An Open-Source Vision-Language-Action Model},
  author={Kim, Moo Jin and Pertsch, Karl and Karamcheti, Siddharth and others},
  booktitle={Conference on Robot Learning (CoRL)},
  year={2024},
  url={https://arxiv.org/abs/2406.09246}
}

@inproceedings{brohan2023rt2,
  title={{RT-2}: Vision-Language-Action Models Transfer Web Knowledge to Robotic Control},
  author={Brohan, Anthony and Brown, Noah and Carbajal, Justice and others},
  booktitle={Conference on Robot Learning (CoRL)},
  year={2023},
  url={https://arxiv.org/abs/2307.15818}
}

@inproceedings{black2024pi0,
  title={{$\pi_0$}: A Vision-Language-Action Flow Model for General Robot Control},
  author={Black, Kevin and Brown, Noah and Driess, Danny and others},
  booktitle={Robotics: Science and Systems (RSS)},
  year={2025},
  url={https://arxiv.org/abs/2410.24164}
}

@inproceedings{madaan2023selfrefine,
  title={Self-Refine: Iterative Refinement with Self-Feedback},
  author={Madaan, Aman and Tandon, Niket and Gupta, Prakhar and others},
  booktitle={Advances in Neural Information Processing Systems},
  year={2023},
  url={https://arxiv.org/abs/2303.17651}
}

@article{chen2026eta,
  title={{ETA}: A New Agentic Paradigm for Embodied Tasks},
  author={Chen, Yitong and Huai, Zezheng and Li, Sixian and Wang, Yubang and Zhang, Haozhe and Zhang, Yifei and Chen, Hechang and Gong, Jingjing and Jiang, Yu-Gang and Qiu, Xipeng},
  journal={arXiv preprint arXiv:2608.03924},
  year={2026},
  url={https://arxiv.org/abs/2608.03924}
}

@article{fu2026capx,
  title={{CaP-X}: A Framework for Benchmarking and Improving Coding Agents for Robot Manipulation},
  author={Fu, Letian and Yu, Justin and El-Refai, Karim and Kou, Ethan and Xue, Haoru and Huang, Huang and Xiao, Wenli and Wang, Guanzhi and Niu, Dantong and Li, Fei-Fei and Shi, Guanya and Wu, Jiajun and Sastry, Shankar and Zhu, Yuke and Goldberg, Ken and Fan, Linxi {Jim}},
  journal={arXiv preprint arXiv:2603.22435},
  year={2026},
  url={https://arxiv.org/abs/2603.22435}
}

@article{lu2026aspire,
  title={{ASPIRE}: Agentic /Skills Discovery for Robotics},
  author={Lu, Runyu and Wu, Yubo and Kou, Ethan and Fu, Letian and Xiao, Wenli and Mandlekar, Ajay and Xu, Yinzhen and Shi, Guanya and Goldberg, Ken and Chen, Ang and Chowdhury, Mosharaf and Zhu, Yuke and Fan, Linxi {Jim} and Wang, Guanzhi},
  journal={arXiv preprint arXiv:2607.00272},
  year={2026},
  url={https://arxiv.org/abs/2607.00272}
}

@article{shi2025maestro,
  title={Maestro: Orchestrating Robotics Modules with Vision-Language Models for Zero-Shot Generalist Robots},
  author={Shi, Junyao and Yang, Rujia and Chao, Kaitian and Wan, Selina Bingqing and Shao, Yifei and Lei, Jiahui and Qian, Jianing and Le, Long and Chaudhari, Pratik and Daniilidis, Kostas and Wen, Chuan and Jayaraman, Dinesh},
  journal={arXiv preprint arXiv:2511.00917},
  year={2025},
  url={https://arxiv.org/abs/2511.00917}
}

@article{huang2026roboharness,
  title={{RoboHarness}: Memory-Driven Orchestration of Heterogeneous Robot Policies for Long-Horizon Planning},
  author={Huang, Jinbang and Hu, Yuanzhao and Li, Zhiyuan and Qi, Ran and Xiao, Yixin and Zhang, Zhanguang and Coates, Mark and Cao, Tongtong and Zhang, Yingxue},
  journal={arXiv preprint arXiv:2607.18060},
  year={2026},
  url={https://arxiv.org/abs/2607.18060}
}

@inproceedings{li2026roboharness,
  title={{RoboHarness}: A Memory-Augmented Policy Harness for Vision-Language-Action Model Robustness via In-Context Adaptation},
  author={Li, Zhuoran and Li, Zhiyang and Zhou, Kaijun and Gu, Jinyu},
  booktitle={IEEE/RSJ International Conference on Intelligent Robots and Systems (IROS)},
  year={2026},
  url={https://arxiv.org/abs/2603.24060}
}

@article{cheng2026gptpolicy,
  title={In-Context Robot Learning with {VLM} Agents},
  author={Cheng, Dongzhou and Yi, Taoran and Fang, Ye and Zhang, Xingwu and Feng, Fan and Li, Yixuan and Zhuang, Gengxiong and Wang, Rongze and Yang, Shuai and Song, Wei and Xue, Weizhi and Wu, Minyan and Gui, Jie and Wang, Jiaqi and Wu, Tong},
  journal={arXiv preprint arXiv:2609.19138},
  year={2026},
  url={https://arxiv.org/abs/2609.19138}
}

@article{sun2025grace,
  title={Executable Analytic Concepts as the Missing Link Between {VLM} Insight and Precise Manipulation},
  author={Sun, Mingyang and Wei, Jiude and He, Qichen and Wang, Donglin and Lu, Cewu and Sun, Jianhua},
  journal={arXiv preprint arXiv:2510.07975},
  year={2025},
  url={https://arxiv.org/abs/2510.07975}
}

@article{ye2025unidomain,
  title={{UniDomain}: Pretraining a Unified {PDDL} Domain from Real-World Demonstrations for Generalizable Robot Task Planning},
  author={Ye, Haoming and Xiao, Yunxiao and Lu, Cewu and Cai, Panpan},
  journal={arXiv preprint arXiv:2507.21545},
  year={2025},
  url={https://arxiv.org/abs/2507.21545}
}

@article{ye2026uniplan,
  title={{UniPlan}: Vision-Language Task Planning for Mobile Manipulation with Unified {PDDL} Formulation},
  author={Ye, Haoming and Xiao, Yunxiao and Lu, Cewu and Cai, Panpan},
  journal={arXiv preprint arXiv:2602.08537},
  year={2026},
  url={https://arxiv.org/abs/2602.08537}
}

@article{fan2026resource2skill,
  title={{RESOURCE2SKILL}: Distilling Executable Agent Skills from Human-Created Multimodal Resources},
  author={Fan, Yijia and Di, Zonglin and Wen, Zimo and Yang, Yifan and Cheng, Mingxi and Dai, Qi and Liu, Bei and Qiu, Kai and Dong, Yue and Li, Ji and Luo, Chong},
  journal={arXiv preprint arXiv:2606.29538},
  year={2026},
  url={https://arxiv.org/abs/2606.29538}
}

@inproceedings{alayrac2022flamingo,
  title={Flamingo: a Visual Language Model for Few-Shot Learning},
  author={Alayrac, Jean-Baptiste and Donahue, Jeff and Luc, Pauline and Miech, Antoine and Barr, Iain and Hasson, Yana and Lenc, Karel and Mensch, Arthur and Millican, Katie and Reynolds, Malcolm and Ring, Roman and Rutherford, Eliza and Cabi, Serkan and Han, Tengda and Gong, Zhitao and Samangooei, Sina and Monteiro, Marianne and Menick, Jacob and Borgeaud, Sebastian and Brock, Andrew and Nematzadeh, Aida and Sharifzadeh, Sahand and Binkowski, Mikolaj and Barreira, Ricardo and Vinyals, Oriol and Zisserman, Andrew and Simonyan, Karen},
  booktitle={Advances in Neural Information Processing Systems (NeurIPS)},
  year={2022},
  url={https://arxiv.org/abs/2204.14198}
}

@inproceedings{liu2023llava,
  title={Visual Instruction Tuning},
  author={Liu, Haotian and Li, Chunyuan and Wu, Qingyang and Lee, Yong Jae},
  booktitle={Advances in Neural Information Processing Systems (NeurIPS)},
  year={2023},
  url={https://arxiv.org/abs/2304.08485}
}

@inproceedings{driess2023palme,
  title={{PaLM-E}: An Embodied Multimodal Language Model},
  author={Driess, Danny and Xia, Fei and Sajjadi, Mehdi S. M. and Lynch, Corey and Chowdhery, Aakanksha and Ichter, Brian and Wahid, Ayzaan and Tompson, Jonathan and Vuong, Quan and Yu, Tianhe and Huang, Wenlong and Chebotar, Yevgen and Sermanet, Pierre and Duckworth, Daniel and Levine, Sergey and Vanhoucke, Vincent and Hausman, Karol and Toussaint, Marc and Greff, Klaus and Zeng, Andy and Mordatch, Igor and Florence, Pete},
  booktitle={International Conference on Machine Learning (ICML)},
  year={2023},
  url={https://arxiv.org/abs/2303.03378}
}

@inproceedings{physical2025pi05,
  title={{$\pi_{0.5}$}: a Vision-Language-Action Model with Open-World Generalization},
  author={{Physical Intelligence} and Black, Kevin and Brown, Noah and Darpinian, James and Dhabalia, Karan and Driess, Danny and Esmail, Adnan and Equi, Michael and Finn, Chelsea and Fusai, Niccolo and Galliker, Manuel Y. and Ghosh, Dibya and Groom, Lachy and Hausman, Karol and Ichter, Brian and Jakubczak, Szymon and Jones, Tim and Ke, Liyiming and LeBlanc, Devin and Levine, Sergey and Li-Bell, Adrian and Mothukuri, Mohith and Nair, Suraj and Pertsch, Karl and Ren, Allen Z. and Shi, Lucy Xiaoyang and Smith, Laura and Springenberg, Jost Tobias and Stachowicz, Kyle and Tanner, James and Vuong, Quan and Walke, Homer and Walling, Anna and Wang, Haohuan and Yu, Lili and Zhilinsky, Ury},
  booktitle={Conference on Robot Learning (CoRL)},
  year={2025},
  url={https://arxiv.org/abs/2504.16054}
}

@article{chen2026evoskillgui,
  title={Reflect, Revise, Reuse: Training-Free Skill Evolution for {GUI} Agents},
  author={Chen, Bofan and Zhang, Boxuan and Tang, Fei and Lu, Zhengxi and Du, Yong and Chen, Tongbo and Lu, Weiming and Xiao, Jun and Zhuang, Yueting and Shen, Yongliang},
  journal={arXiv preprint arXiv:2609.17653},
  year={2026},
  url={https://arxiv.org/abs/2609.17653}
}

@article{xu2026omniharness,
  title={{OmniHarness}: Harnessing Generalizable Visual Generation via Symbolic Policy Learning},
  author={Xu, Xu and Liu, Jinxiu and Qiao, Zhangbo and Lu, Jiaxing and Zhang, Xiangyu and Gu, Yubin and Ning, Fangwei and Shi, Yan},
  journal={arXiv preprint arXiv:2609.16057},
  year={2026},
  url={https://arxiv.org/abs/2609.16057}
}

@article{ju2026embodiskill,
  title={{EmbodiSkill}: Skill-Aware Reflection for Self-Evolving Embodied Agents},
  author={Ju, Ruofei and Wang, Xinrui and Ding, Xin and Yang, Yifan and Wu, Hao and Jiang, Shiqi and Zhang, Qianxi and Wen, Hao and Li, Xiangyu and Wang, Weijun and Li, Kun and Liu, Yunxin and Dai, Haipeng and Wang, Wei and Cao, Ting},
  journal={arXiv preprint arXiv:2605.10332},
  year={2026},
  url={https://arxiv.org/abs/2605.10332}
}

@article{zhou2026zerowam,
  title={{Zero-WAM}: In-Context World-Action Modeling from Human Videos for Open-Ended Task Generalization},
  author={Zhou, Jiaming and Zhang, Qihang and Xu, Gangwei and Fan, Cunxin and Zhao, Yujie and Wang, Ruilin and Luo, Yiming and Yang, Shuai and Zhu, Xing and Shen, Yujun and Liang, Junwei and Xu, Yinghao},
  journal={arXiv preprint arXiv:2608.26103},
  year={2026},
  url={https://arxiv.org/abs/2608.26103}
}

@article{gemini2024gemini15,
  title={Gemini 1.5: Unlocking multimodal understanding across millions of tokens of context},
  author={{Gemini Team} and others},
  journal={arXiv preprint arXiv:2403.05530},
  year={2024},
  url={https://arxiv.org/abs/2403.05530}
}

@article{gemini2025robotics15,
  title={Gemini Robotics 1.5: Pushing the Frontier of Generalist Robots with Advanced Embodied Reasoning, Thinking, and Motion Transfer},
  author={{Gemini Robotics Team} and others},
  journal={arXiv preprint arXiv:2510.03342},
  year={2025},
  url={https://arxiv.org/abs/2510.03342}
}

@inproceedings{xiao2026enpire,
  title={{ENPIRE}: Agentic Robot Policy Self-Improvement in the Real World},
  author={Xiao, Wenli and Xie, Jia and Zhang, Tonghe and Lin, Haotian and Fu, Letian and Xue, Haoru and Lu, Jalen and Yang, Yi and Dai, Cunxi and Wang, Zi and Wu, Jimmy and Wang, Guanzhi and Sastry, S. Shankar and Goldberg, Ken and Fan, Linxi and Zhu, Yuke and Shi, Guanya},
  booktitle={Conference on Robot Learning (CoRL)},
  year={2026},
  url={https://arxiv.org/abs/2606.19980}
}

@inproceedings{huang2023voxposer,
  title={{VoxPoser}: Composable {3D} Value Maps for Robotic Manipulation with Language Models},
  author={Huang, Wenlong and Wang, Chen and Zhang, Ruohan and Li, Yunzhu and Wu, Jiajun and Li, Fei-Fei},
  booktitle={Conference on Robot Learning (CoRL)},
  year={2023},
  url={https://arxiv.org/abs/2307.05973}
}

@inproceedings{huang2024rekep,
  title={{ReKep}: Spatio-Temporal Reasoning of Relational Keypoint Constraints for Robotic Manipulation},
  author={Huang, Wenlong and Wang, Chen and Li, Yunzhu and Zhang, Ruohan and Li, Fei-Fei},
  booktitle={Conference on Robot Learning (CoRL)},
  year={2024},
  url={https://arxiv.org/abs/2409.01652}
}

@inproceedings{huang2022language,
  title={Language Models as Zero-Shot Planners: Extracting Actionable Knowledge for Embodied Agents},
  author={Huang, Wenlong and Abbeel, Pieter and Pathak, Deepak and Mordatch, Igor},
  booktitle={International Conference on Machine Learning (ICML)},
  year={2022},
  url={https://arxiv.org/abs/2201.07207}
}

@inproceedings{singh2023progprompt,
  title={{ProgPrompt}: Generating Situated Robot Task Plans using Large Language Models},
  author={Singh, Ishika and Blukis, Valts and Mousavian, Arsalan and Goyal, Ankit and Xu, Danfei and Tremblay, Jonathan and Fox, Dieter and Thomason, Jesse and Garg, Animesh},
  booktitle={IEEE International Conference on Robotics and Automation (ICRA)},
  year={2023},
  url={https://arxiv.org/abs/2209.11302}
}

@inproceedings{liu2023reflect,
  title={{REFLECT}: Summarizing Robot Experiences for Failure Explanation and Correction},
  author={Liu, Zeyi and Bahety, Arpit and Song, Shuran},
  booktitle={Conference on Robot Learning (CoRL)},
  year={2023},
  url={https://arxiv.org/abs/2306.15724}
}

@inproceedings{guo2024doremi,
  title={{DoReMi}: Grounding Language Model by Detecting and Recovering from Plan-Execution Misalignment},
  author={Guo, Yanjiang and Wang, Yen-Jen and Zha, Lihan and Chen, Jianyu},
  booktitle={IEEE/RSJ International Conference on Intelligent Robots and Systems (IROS)},
  year={2024},
  url={https://arxiv.org/abs/2307.00329}
}

@inproceedings{zhou2025codeasmonitor,
  title={{Code-as-Monitor}: Constraint-aware Visual Programming for Reactive and Proactive Robotic Failure Detection},
  author={Zhou, Enshen and Su, Qi and Chi, Cheng and Zhang, Zhizheng and Wang, Zhongyuan and Huang, Tiejun and Sheng, Lu and Wang, He},
  booktitle={IEEE/CVF Conference on Computer Vision and Pattern Recognition (CVPR)},
  year={2025},
  url={https://arxiv.org/abs/2412.04455}
}

@inproceedings{duan2025aha,
  title={{AHA}: A Vision-Language-Model for Detecting and Reasoning Over Failures in Robotic Manipulation},
  author={Duan, Jiafei and Pumacay, Wilbert and Kumar, Nishanth and Wang, Yi Ru and Tian, Shulin and Yuan, Wentao and Krishna, Ranjay and Fox, Dieter and Mandlekar, Ajay and Guo, Yijie},
  booktitle={International Conference on Learning Representations (ICLR)},
  year={2025},
  url={https://arxiv.org/abs/2410.00371}
}

@inproceedings{zhao2023act,
  title={Learning Fine-Grained Bimanual Manipulation with Low-Cost Hardware},
  author={Zhao, Tony Z. and Kumar, Vikash and Levine, Sergey and Finn, Chelsea},
  booktitle={Robotics: Science and Systems (RSS)},
  year={2023},
  url={https://arxiv.org/abs/2304.13705}
}
\normalem

\clearpage
\appendix
\section{TSR Formulation and Analysis}
\label{sec:tsr-analysis}
\paragraph{Task--skill graph and scoped patches.}
At iteration $t$, the released hierarchy is represented by
$G_t=(V_t,E_t,\{c_v\}_{v\in V_t})$, where $V_t$ contains task and skill nodes,
$E_t$ contains possible call edges, and $c_v$ is a node's implementation or
guidance. The execution record $\tau_t$ includes the goal, observations, and
tool trace. Diagnosis identifies the affected scope $S_t\subseteq V_t$;
$G_t|_{S_t}$ contains its nodes and internal calls, while $\partial S_t$
contains the interfaces connecting it to the unchanged hierarchy.

The revision procedure $\mathcal R$ proposes a graph and code patch
$\Delta_t$ with $\operatorname{supp}(\Delta_t)\subseteq S_t$. The support
identifies existing nodes whose implementation, guidance, or outgoing calls
change. New helper nodes may be attached within that scope. The operator
$\oplus$ applies both code edits and branch expansion. If diagnosis
implicates a caller or interface outside $S_t$, the scope expands before
the next proposal.

\paragraph{History and validation.}
The case collection in Eq.~\ref{eq:tsr-refinement} is
$\mathcal H_t=\mathcal D_t^{-}\cup\mathcal D_t^{+}(S_t)$, comprising recent
failures that motivate the edit and selected successful cases from affected
callers. Review and functional tests determine whether $\mathcal V$ returns
the proposed patch or an empty patch. Applying an empty patch leaves the
released graph unchanged; accepted revisions become available to subsequent
execution. Shared skills require checking the relevant caller paths even
when only one implementation changes. The next iteration uses feedback
from the currently released version.

\paragraph{Dependency-local effects.}
Equation~\ref{eq:tsr-refinement} separates the scope of an edit from the
scope of its possible effects. Let $r_q$ be the entry node for task $q$ and
let $\operatorname{Reach}_{G_t}(r_q)$ include that node and all nodes
reachable through possible calls. The affected task set is
$\mathcal A_t=\{q:\operatorname{Reach}_{G_t}(r_q)\cap S_t\ne\emptyset\}$.
This uses possible dependencies, not only calls observed in one rollout:
a shared base skill may affect several tasks under different observations.

Consider a fixed task distribution $p(q)$, and let $J_q(G)$ be the expected
failure rate for task $q$ under graph $G$. Assume the graph captures all
behavioral dependencies, the patch leaves interfaces and implementations
outside its scope unchanged, and evaluation uses the same reset and
exogenous-randomness distributions. In particular, edits must not alter
untracked shared state, global prompts, or tool configurations. Under these
conditions, tasks outside $\mathcal A_t$ cannot encounter a changed
implementation. Their execution distributions are unchanged, so their
contributions cancel in the aggregate difference:
\begin{equation}
\begin{aligned}
J(G_{t+1})-J(G_t)
 &=\sum_{q\in\mathcal A_t}p(q)
     \bigl[J_q(G_{t+1})-J_q(G_t)\bigr],\\
\left|J(G_{t+1})-J(G_t)\right|
 &\leq\sum_{q\in\mathcal A_t}p(q),
 \qquad J(G)=\sum_q p(q)J_q(G).
\end{aligned}
\label{eq:tsr-dependency-effect}
\end{equation}
The bound follows because each failure rate lies in $[0,1]$. Its magnitude
scales with the probability mass of affected tasks and covers both
improvements and regressions. A task-specific branch can have a small
affected set, whereas a widely shared base skill can have a large one.
This motivates selecting regression cases by dependency coverage rather
than by the number of edited functions. If a revision changes shared state
or an interface, its dependency scope must be expanded before this analysis
applies. Development cases guide revision, while held-out tasks measure
generalization.

\paragraph{Token accounting.}
Construction cost includes skill generation, diagnosis, review, and
validation. Execution cost includes all model calls within the executing
agent, including model-backed perception, monitoring, and fallback;
separately measured planning and review are excluded from this execution
subtotal. Unmetered calls are reported separately. Comparisons require
matched task distributions and must include failed executions, since a
shorter failed attempt can consume fewer tokens without improving efficiency.
Task success, runtime, and the frequency of fallback to online reasoning
provide context for interpreting any reduction in execution tokens.

\paragraph{Execution and outcome interfaces.}
Agents communicate through plans, structured skill inputs and outputs, and
execution records. Physical deployment uses human guidance and safety
supervision; simulation can automate resets and outcome checks. Planner,
Engineer, and Reviewer receive observations and tool feedback, while hidden
simulator state and success-condition internals remain outside their
context. The simulator scores the task separately after execution.

\paragraph{Operational details.}
Each skill has a descriptor that declares its arguments, typed return
fields, applicability conditions, and a validation harness (simulator task,
seeds, arguments, and any setup skill). The return fields report whether the
intended effect was verified; for example, a placement skill returns whether
the target was reached, the gripper opened, and the release was verified,
and it does not open the gripper unless its reach and hold checks pass. An
atomic task counts as complete only when its postcondition holds on fresh
observations taken after the last action. A candidate revision is staged in
an isolated copy of the library and reviewed by the Manager. If it changes a
base skill, the harness of that skill runs in the simulator; the complete
atomic task is then re-run on its development seeds. Release requires
simulator success on at least two declared seeds, restoration of the
unmodified source after the trial, and Manager approval with no outstanding
changes; a candidate that fails cannot be resubmitted unchanged. In
simulation, $\mathcal H_t$ consists of these harness and development runs;
on the robot, it consists of the four most recent eligible runs, replayed
offline (Appendix~\ref{sec:physical-revisions}). A compound skill is
proposed once a task has at least three successful executions whose tool
sequences share at least 60\% of their longest common subsequence and a
common skeleton of at least three steps. The generated skill replays the
skeleton and recomputes every argument from perception at run time, never
storing coordinates from an episode, and it passes the same review and
validation as any other revision.

\section{Physical Evaluation Setting \& Protocol}
\label{sec:mobile-real-method}

\paragraph{Platform and observation interface.}
The physical experiments use two robot platforms. The mobile cleanup
case study uses a WheelSingleArm M1 comprising an Athena Pro Max mobile
base, a RealMan arm, and a Zhiyuan gripper. An in-hand camera and a
chassis-mounted camera provide RGB-D observations. The agent can access
arm-joint and end-effector states, gripper state, chassis pose, and
navigation and docking status. Camera-to-arm calibration was performed
before the experiments. All mobile development runs used the same
prebuilt navigation map.

The dual-arm setup comprises two AgileX PiPER robotic arms with
standard 100-mm PiPER grippers. An Intel RealSense D435i facing the
workspace and an Intel RealSense D405 mounted on each wrist provide
RGB-D observations. All three camera streams are available to the
agent, together with joint states, end-effector poses, and gripper
states for both arms. Camera-to-arm calibration was performed before
the experiments.

\paragraph{Mobile task distribution and development protocol.}
The mobile cleanup development record spans 104 runs. Each run starts
from a common docking pose. Before execution, on-site personnel
restore the scene to a common base layout while varying the number,
identities, positions, and orientations of the objects, as well as
the positions of their destination receptacles. The task manifest
specifies the objects and destinations for that run and remains fixed
during its execution. The robot is expected to place the specified
objects in their designated receptacles and return to the docking
pose.

RoboRSI may revise its skills between runs, while the skill version
remains fixed within each run. Revisions are made when needed rather
than after every run. Candidate revisions are checked against
retained, reviewed execution records before use in subsequent runs;
earlier tasks are not physically repeated for this check. Thus, the
104 runs trace development of an evolving system across varying
task manifests, rather than repeated trials of a frozen policy on
an identical task.

\paragraph{Outcome annotation and figure construction.}
A placement is counted when on-site personnel observe the object
enter its designated receptacle. For every run selected for
Figure~\ref{fig:physical-four-item-progress}, each counted placement
was confirmed in this way. The figure shows ten runs selected
retrospectively to represent early partial execution, multi-object
placement, setbacks during development, and later completed tasks.
Selection was made after inspection of the development record, not
according to a prespecified sampling rule.

The plotted value is the number of confirmed placements among the
first four items handled in a run. Four items provide a common display
scale; a run with fewer than four items in its manifest is not treated
as a four-item trial. Marker appearance indicates the audited task
outcome. Because object sets and reset configurations vary across
runs, the plotted values describe placement progress during
development rather than success rates on a fixed task.

\paragraph{Mobile execution records.}
The recorded demonstration comprises two cleanup scenes with five
and four target objects, respectively
(Figure~\ref{fig:real-world}). On-site personnel confirmed that
every object was placed in its designated receptacle. Between
scenes, an operator rearranged the objects and receptacles; the
skill code, prompts, and policy configuration remained unchanged.
Neither cleanup execution required manual takeover or correction.

A separate physical evaluation attempt illustrates a placement
failure: the robot grasped a bath pouf but released it outside the
target bucket. The execution trace exposed a mismatch between the
release check and the required final state, directing a subsequent
revision of the placement geometry. The revision was checked using
retained observations, but was not evaluated in a matched
post-repair physical trial. This failure case is reported separately
from the two-scene demonstration and the 104-run development trace.
\section{Supplementary Experiments}
\label{sec:result-inventory}
This appendix reports the failure classification, exploratory development, and small-scale policy learning.

\subsection{Failure Classification}
\label{sec:failure-taxonomy}
Each failed episode receives one category from deterministic rules over its
public tool results and the simulator's terminal outcome. Explicit fields
such as an empty grasp, a dropped object, a rejected release, or an
unreachable target determine grasp, placement/release, and
motion/reachability failures; for complete traces, the last explicit failure
step is used. An episode in which the agent declares success while the
simulator reports failure is a premature completion, and an episode that
ends at the interaction limit is budget exhausted. Episodes whose traces
support none of these rules are unlabeled. Categories are counted separately
for each evaluation panel; the analysis covers 5,008 failed episodes.
Completion after a failure (Figure~\ref{fig:failure-analysis}c) is computed
on the (task, seed) pairs for which every method has a complete trace and
encounters at least one execution failure, 209 on LIBERO-PRO and 258 on
LIBERO-Plus; it is the fraction of these episodes that still end in
simulator success.

\subsection{Exploratory Development}
Early LIBERO-Plus runs also allowed capability updates on the target
benchmark. Their configurations allowed repeated attempts and target-side updates,
whereas the main LIBERO-Plus transfer panel keeps source capabilities fixed.

\subsection{Learned Policies from Execution Data}
\label{sec:learned-policy-experiments}
Demonstrations are successful RoboRSI executions on fixed collection
layouts: ten for \texttt{click\_bell}, yielding 2,161 training samples, and
seven for \texttt{click\_alarmclock}. Both action representations use the
same ACT architecture, with a ResNet-18 visual backbone trained from
scratch, a transformer width of 256, action chunks of 100 steps of which 25
are executed per inference, a KL weight of 1, and 10,000 training steps.
The absolute version predicts joint position commands directly. The
chunk-relative version predicts joint positions relative to the arm
configuration at the start of each chunk, with gripper and velocity outputs
unchanged. Final checkpoints run through a standalone inference server
without online agent assistance, with a budget of 2,000 control ticks per
episode. Evaluation covers the ten training layouts of \texttt{click\_bell}
and the ten collection layouts of \texttt{click\_alarmclock}, seven of which
supplied training demonstrations.

\section{Case Studies: Skill Code and Failure-Driven Revision}
\label{sec:cases}
We examine an execution failure and a local skill revision on
the same development task and initial states. Separate cases then document skill
publication and review-driven code revision.

\lstdefinestyle{casecode}{language=Python,basicstyle=\ttfamily\scriptsize,
  keywordstyle=\bfseries,commentstyle=\itshape,stringstyle=\ttfamily,
  breaklines=true,breakatwhitespace=false,showstringspaces=false,
  columns=fullflexible,keepspaces=true,frame=none,aboveskip=6pt,belowskip=6pt}

\subsection{Execution Repair: Removing a Premature Target Rejection}
\label{sec:case-execution-repair}
This case follows an earlier revision of
\texttt{guarded\_compact\_feature\_tap} on \texttt{click\_bell}. Both
candidates were reviewed by GPT-5.6-SOL Manager calls and evaluated on the
same three declared initial states, invocation arguments, and independent
simulator-success criterion. Each version ran once per initial state. The
original candidate succeeded in one of the three trials (1/3); the revised
candidate succeeded in all three (3/3), with no crashes. It met the
predeclared requirement of at least two successes and was then published.
These six runs form the development comparison below.

\paragraph{Failure and responsible node.}
In the first failed trial, the original skill observed a bell bounding box
$[249,185,306,240]$ in a 320$\times$240 image. Its lower edge exceeded the
hard-coded threshold of 237, so the target-association check rejected the
object before issuing a tap. The task consequently failed. The responsible
code lay in the skill's association checks following
\texttt{get\_object\_bbox}; the bell-pressing goal and tool interface did
not need to change. The original size and boundary checks were:
\begin{lstlisting}[style=casecode]
if width < 8 or height < 8 or width > 100 or height > 100:
    return fail("object association is not within fixed compactness bounds")
if x0 < 2 or y0 < 2 or x1 > 317 or y1 > 237:
    return fail("object association touches the fixed head-camera border guard")
\end{lstlisting}

\paragraph{Local revision and observed execution.}
After receiving the failed runs' public tool traces, the Manager generated
a revision that allows bounded lower-edge truncation for a compact object.
The revised checks below retain size limits and reject extensive truncation;
feature-in-box, depth, and arm-state checks remain elsewhere in the skill,
and the revision also adds motion-endpoint checks.
\begin{lstlisting}[style=casecode]
if width < 8 or height < 8 or width > 100 or height > 100:
    return fail("object association is outside fixed compactness bounds")
if x0 < 2 or y0 < 2 or x1 > 318 or y1 > 240:
    return fail("object association violates fixed image guards")
if y1 == 240 and height > 70:
    return fail("lower-edge truncation is too extensive")
\end{lstlisting}

\begin{figure}[!htbp]
\centering
\includegraphics[width=\linewidth]{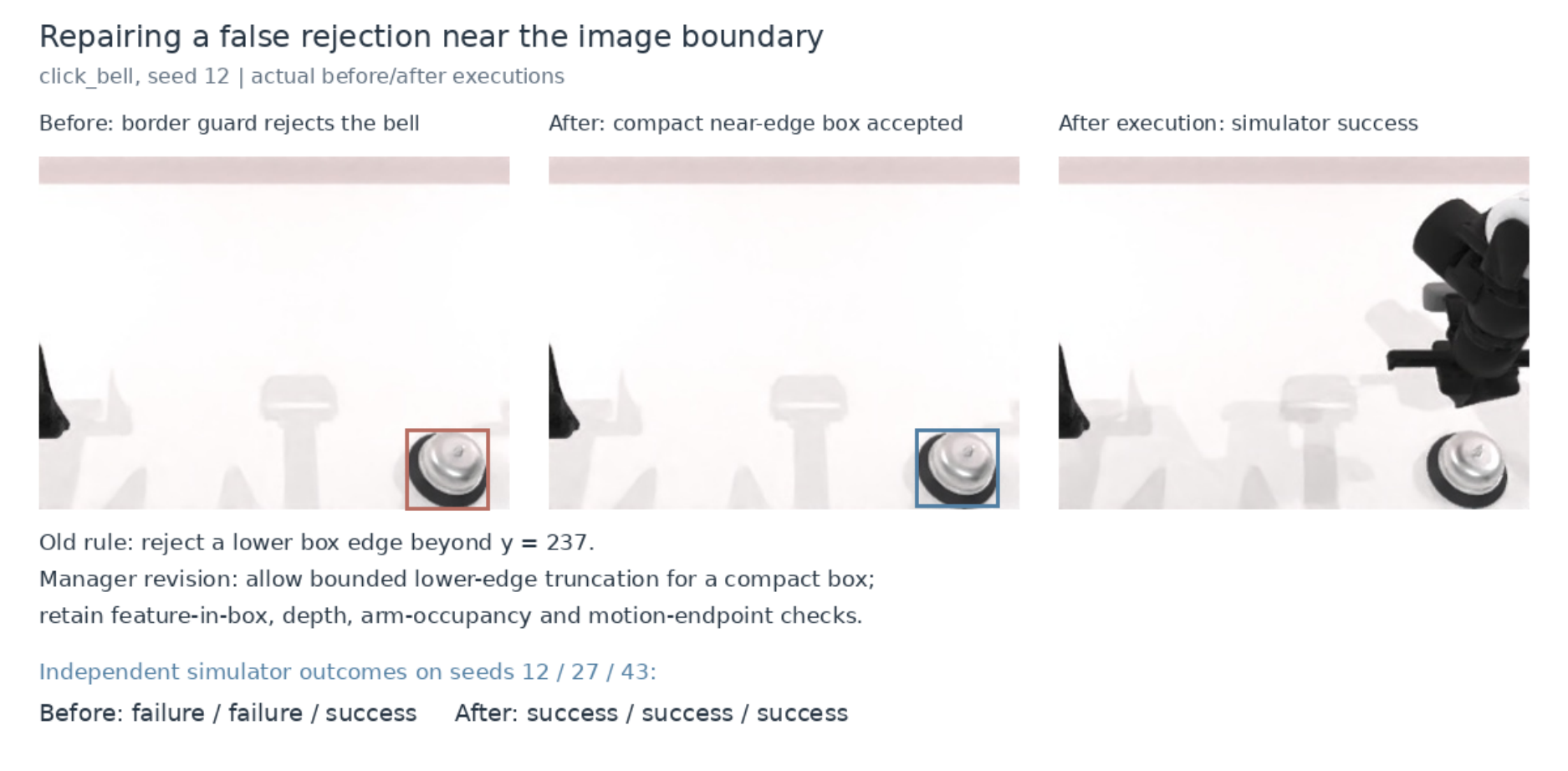}
\caption{A local skill revision enables execution in the
bell-pressing case. Left: the original target-association check rejects the
near-boundary box before action. Middle: the revised check accepts the
compact target. Right: the post-execution observation from the successful
revised run. Boxes mark the detected targets; the simulator supplies the
task outcomes.}
\label{fig:tsr-execution-repair}
\end{figure}

In the revised run from the same initial state, the observed box was
$[249,185,306,239]$ and the localized feature was $(257,195)$.
The skill accepted the target, called \texttt{move\_to\_pixel} with
\texttt{action=tap}, and received the hover--close--press--retreat execution
report. The independent simulator verdict was success.
Figure~\ref{fig:tsr-execution-repair} shows the corresponding observations.
Replaying the old code on this same recorded post-revision observation
still rejects the box because $239>237$; the revised code replays all 22
recorded public calls, including the tap. The replay uses stored tool
outputs to compare the two implementations on the same observation.

\paragraph{Evaluation setting.}
A supervisor supplied the complete tool arguments, outputs, and failed
trace to the Manager and organized the two validation runs. The Manager
revised the code, followed by a separate review. The patch updated the
boundary checks and recovery logic; in the second trial, both versions
attempted a tap.
The native tools, dispatch, and harness code were unchanged. Runs used
different GPUs and slightly different initial images, including a one-pixel
difference in the box boundary of the first trial. Process limits were 300 and 900
seconds, with actual durations of approximately 231 and 297 seconds.

\subsection{Generated Compound Skill: Validation, Publication, and Invocation}
\label{sec:case-consolidated}
A later, distinct \texttt{exact\_waypoint\_feature\_tap} skill provides a concrete
example of executable capability accumulation. A GPT-5.6-SOL Manager call
generated the revised code, and a separate Manager call reviewed it.
The descriptor exposes four inputs: arm, object description, functional
feature description, and camera. The implementation obtains a fresh image,
localizes the object and its functional point, recovers depth, checks motion
endpoints, executes a tap, and checks retreat and the terminal state.
Its structured outputs describe execution, recovery, and failure reasons.
The simulator evaluates task completion.

The following source excerpt shows the fixed execution sequence after
perception. Here \texttt{xyz} is the currently observed feature position,
\texttt{u,v} its image coordinates, and \texttt{screen\_tcp} checks the
candidate endpoint. Input validation, perception, and subsequent terminal
checks are omitted from this excerpt. The code preserves the procedure,
while object association and action targets are recomputed on each call.
\begin{lstlisting}[style=casecode]
top_down = [0.5, -0.5, 0.5, 0.5]
approach_tcp = [xyz[0], xyz[1], xyz[2] + 0.10] + top_down
press_tcp = [xyz[0], xyz[1], xyz[2] - 0.005] + top_down

if screen_tcp(approach_tcp) is None:
    return fail("exact tap approach/retreat endpoint is unsafe or unknown")
if screen_tcp(press_tcp) is None:
    return fail("exact tap press endpoint is unsafe or unknown")

tap_result, _ = _dispatch_tool(
    state,
    "move_to_pixel",
    {
        "arm": arm,
        "u": u,
        "v": v,
        "action": "tap",
        "height_above_m": 0.0,
        "camera": "head_camera",
    },
)
tapped = tap_result.get("ok") is True

\end{lstlisting}

\paragraph{Validation and publication.}
The predeclared development check required independent simulator success
on at least two of three \texttt{click\_bell} trials. Two of the three
trials succeeded, and no run crashed. The 2/3 result passed that check,
after which the reviewed implementation was published as a versioned skill.
This later skill uses its own three-trial development panel.

\begin{figure}[!htbp]
\centering
\includegraphics[width=\linewidth]{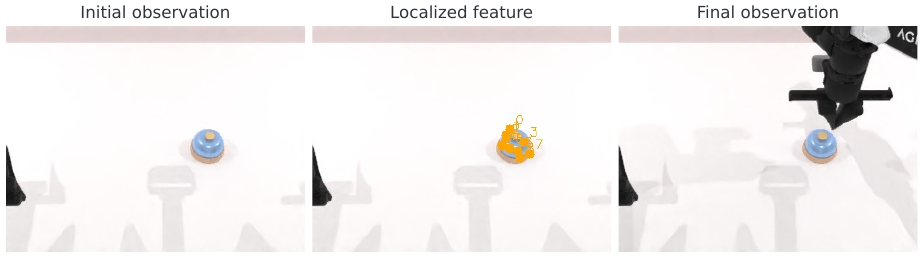}
\caption{Observations from the published tap skill in a successful
\texttt{click\_bell} development trial: the initial view, feature
localization, and post-execution view. The outcome is scored by the
simulator; images are 320$\times$240 pixels.}
\label{fig:skill-publication-case}
\end{figure}

Figure~\ref{fig:skill-publication-case} shows the observations from a
successful development trial.

\subsection{Before/After Code: Review-Driven Terminal Checks}
\label{sec:case-code-repair}
The earlier candidate was rejected during Manager review. Its recovery
expression short-circuited: when a movement or pose check failed, the later
occupancy and clearance queries could be skipped. Elsewhere, clearance
could be queried with empty-arm assumptions after occupancy was not
confirmed. The old recovery condition was:
\begin{lstlisting}[style=casecode]
recovered = (
    recovery_move.get("ok") is True
    and recovery_measure.get("ok") is True
    and recovery_measure.get("pose_matches") is True
    and both_empty()
    and current_clear()
)
\end{lstlisting}

The model-generated revision evaluates terminal occupancy after each
recovery attempt, then checks clearance only if both arms are confirmed
empty. It uses these recorded checks when deciding whether recovery was
established. The helper and revised condition are shown below; intervening
motion calls are omitted.
\begin{lstlisting}[style=casecode]
def terminal_empty_clear():
    empty = both_empty()
    if not empty:
        return False, False
    return True, empty_current_clear()

recovery_empty, recovery_clear = terminal_empty_clear()
recovered = (
    recovery_move.get("ok") is True
    and recovery_measure.get("ok") is True
    and recovery_measure.get("pose_matches") is True
    and recovery_empty
    and recovery_clear
)
\end{lstlisting}

A separate Manager review accepted this version, which passed the 2/3
simulator development check before publication. The previous candidate
stopped at code review and had no simulator runs. The full patch also
changes diagnostics and removes an optional visual-difference check.
An operator separately maintained the camera, low-level interfaces, and
supervisor. The code excerpts show the terminal-check revision.

\subsection{Physical Skill Revisions in Rounds 25 and 33}
\label{sec:physical-revisions}
Both records come from the household cleanup task on the mobile
manipulator (Section~\ref{sec:real-world}) and are shown in
Figure~\ref{fig:tsr-evolution}b. Each revision was admitted with
its trigger run, the four most recent eligible runs as a replay buffer, and
the changed files; live iteration was paused until validation finished.

\paragraph{Round 25: placement observation and release.}
\emph{Execution.} The robot held a food wrapper in front of the correct
bag-lined trash bin, which remained visually distinct from a nearby
light-blue bucket. The compiled observation move reported a 0.052\,m and
17.1$^\circ$ tolerance miss, although the current joint and end-effector
state and the third-person video showed the arm at the requested pose.
The unrestricted fallback repeated opening estimation from the same view,
invoked Arm Explore with an invalid self-inspection target, and exhausted
28 merged tool calls. A workspace probe had already stored a complete
three-step release sequence in runtime memory, but the model passed an
incomplete copy to \texttt{place\_object}. Both calls failed closed before
motion; the gripper never opened, and the robot returned to its docked home
pose with the item still held.
\emph{Diagnosis.} The earliest failing node was completion of the placement
observation motion. Its false negative expanded into duplicated geometry
estimation and an incomplete release plan.
\emph{Revision.} (i) A failed observation move receives one read-only pose
recheck, and a physically reached pose continues without replaying the
placement chain. (ii) An unreached pose returns a non-releasable checkpoint
to the Planner, with Engineer fallback disabled. (iii) Probing and release
above a container require opening geometry from the current observation
rather than semantic rim identity alone. (iv) \texttt{place\_object}
replaces an incomplete model-written sequence with the complete sequence
held in runtime destination memory.
\emph{Validation.} Five new regression tests, 172 placement tests, 863
manager tests, and 97 real-robot interface tests passed. The next run
required a fresh human reset of the scene.

\paragraph{Round 33: destination identity and navigation handoff.}
\emph{Execution.} A single sweep over five headings recorded all eight
expected scene entities. The first item was released and its detachment
confirmed, but the run ended after about 25 minutes with the second item
still held. Two owners failed. The inventory producer chose an isolated
but edge-cropped view of the destination bucket although the same sweep
contained a complete view, and the consumer then treated the blue color
ratio as a mandatory check and rejected the producer-certified destination.
Earlier, a navigation to the item had been attempted but not completed and
produced no post-navigation image; this state was routed to Engineer
fallback, which repeated status, power, and camera queries and exhausted
the merged budget before a single compiled retry succeeded.
\emph{Revision.} Inventory selection prefers complete, non-edge
observations; query color becomes advisory during materialization because
the producer owns semantic identity, while current-frame navigation and
manipulation checks remain in force; and incomplete navigation without a
post-navigation image returns a typed checkpoint to the Planner instead of
entering fallback. The revision adds no new object or destination branch,
no episode coordinates, no historical pose gate, and no change to global
color thresholds.
\emph{Validation.} An offline replay of the trigger run selected the
complete bucket view, materialized the destination, and routed the
incomplete navigation to the Planner. The replay buffer contained four
consecutive blocked runs with different terminal failures: a true
destination rejected by a score veto, a defensive post-release check,
exhausted placement regions, and the color veto above. Twenty-eight focused
tests, 866 manager tests, and 97 real-robot interface tests passed. Resets
before and after the run were planned; no unplanned human prompt occurred
during execution.

\subsection{Corrections That Required Human Knowledge}
\label{sec:physical-cases}
\begin{figure}[!htbp]
\centering
\includegraphics[width=\linewidth]{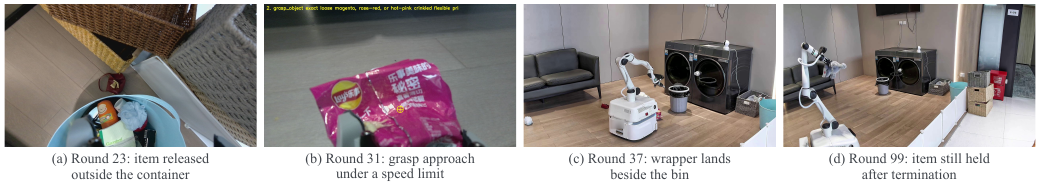}
\caption{\textbf{Physical corrections that required human knowledge.}
(a) Resetting the scene after a confirmed misplacement; (b) a safe grasp
speed specified by the operator; (c) a wrapper released outside the bin;
(d) holding protection after termination.}
\label{fig:human-knowledge}
\end{figure}

Four of the 53 classified revision triggers in the physical development
(Section~\ref{sec:real-world}) required knowledge that the system could not
obtain from its own records (Figure~\ref{fig:human-knowledge}).
In Round 23, a released item landed outside the intended container. The
operator specified that a confirmed misplacement contaminates the scene and
must be reset by a person, rather than being treated as a new source object
for the robot to recover.
In Round 31, the grasp approach had to respect a default linear speed of
0.02\,m/s that the validated motion sequence did not yet consume; the
operator supplied and confirmed this safety limit.
In Round 37, a wrapper fell beside the bin after release, and the operator
judged the placement a failure that required a scene reset, not a success
or a step for automatic recovery.
In Round 99, the robot still held an item when the task stopped, and the
item was later lost while the robot waited at its dock. The operator
specified that holding protection must inherit the reviewed grip width and
force and, when the hold is lost, freeze motion and request confirmation
instead of moving or releasing.
Each correction is kept as a reusable contract, checkpoint, or test on the
responsible skill, so the same prompt is not needed again.

\end{document}